\documentclass[acmsmall]{acmart}

\usepackage{amsmath,amsfonts,amssymb}
\usepackage{graphicx}
\usepackage{setspace}
\usepackage{algorithm}
\usepackage{algorithmic}
\usepackage{multirow}
\usepackage{color}
\usepackage{longtable}
\usepackage{tabularx}
\usepackage{rotating}

\AtBeginDocument{%
  }

\acmJournal{TOMM}

\begin{document}

\title{Learning Semantic-Aware Representation in Visual-Language Models for Multi-Label Recognition with Partial Labels}

\renewcommand{\thefootnote}{}
\footnotetext{Haoxian Ruan and Zhihua Xu share first authorships. Tianshui Chen is the Corresponding author.}
\author{Haoxian Ruan}
\email{HowxRuan@mail2.gdut.edu.cn}
\affiliation{%
  \institution{Guangdong University of Technology}
  \city{Guangzhou}
  \state{Guangdong}
  \country{China}
}

\author{Zhihua Xu}
\email{zihua@mail2.gdut.edu.cn}
\affiliation{%
  \institution{Guangdong University of Technology}
  \city{Guangzhou}
  \state{Guangdong}
  \country{China}
}

\author{Zhijing Yang}
\affiliation{%
  \institution{Guangdong University of Technology}
  \city{Guangzhou}
  \state{Guangdong}
  \country{China}
}
\email{yzhj@gdut.edu.cn}

\author{Yongyi Lu}
\affiliation{%
  \institution{Guangdong University of Technology}
  \city{Guangzhou}
  \state{Guangdong}
  \country{China}
  }
\email{yylu@gdut.edu.cn}

\author{Jinghui Qin}
\affiliation{%
  \institution{Guangdong University of Technology}
  \city{Guangzhou}
  \state{Guangdong}
  \country{China}
}
\email{qinjinghui@gdut.edu.cn}

\author{Tianshui Chen*}
\affiliation{%
  \institution{Guangdong University of Technology}
  \city{Guangzhou}
  \state{Guangdong}
  \country{China}
 }
 \email{tianshuichen@gmail.com}

\renewcommand{\shortauthors}{Ruan et al.}

\begin{abstract}
Multi-label recognition with partial labels (MLR-PL), in which only some labels are known while others are unknown for each image, is a practical task in computer vision, since collecting large-scale and complete multi-label datasets is difficult in real application scenarios. Recently, vision language models (e.g. CLIP) have demonstrated impressive transferability to downstream tasks in data limited or label limited settings. However, current CLIP-based methods suffer from semantic confusion in MLR task due to the lack of fine-grained information in the single global visual and textual representation for all categories. In this work, we address this problem by introducing a semantic decoupling module and a category-specific prompt optimization method in CLIP-based framework. Specifically, the semantic decoupling module following the visual encoder learns category-specific feature maps by utilizing the semantic-guided spatial attention mechanism. Moreover, the category-specific prompt optimization method is introduced to learn text representations aligned with category semantics. Therefore, the prediction of each category is independent, which alleviate the semantic confusion problem. Extensive experiments on Microsoft COCO 2014 and Pascal VOC 2007 datasets demonstrate that the proposed framework significantly outperforms current state-of-art methods with a simpler model structure. Additionally, visual analysis shows that our method effectively separates information from different categories and achieves better performance compared to CLIP-based baseline method.
\end{abstract}





\maketitle

\section{Introduction}
\label{sect:intro}  

\begin{figure*}[!t] 
	\centering 
	\includegraphics[width=0.9\linewidth]{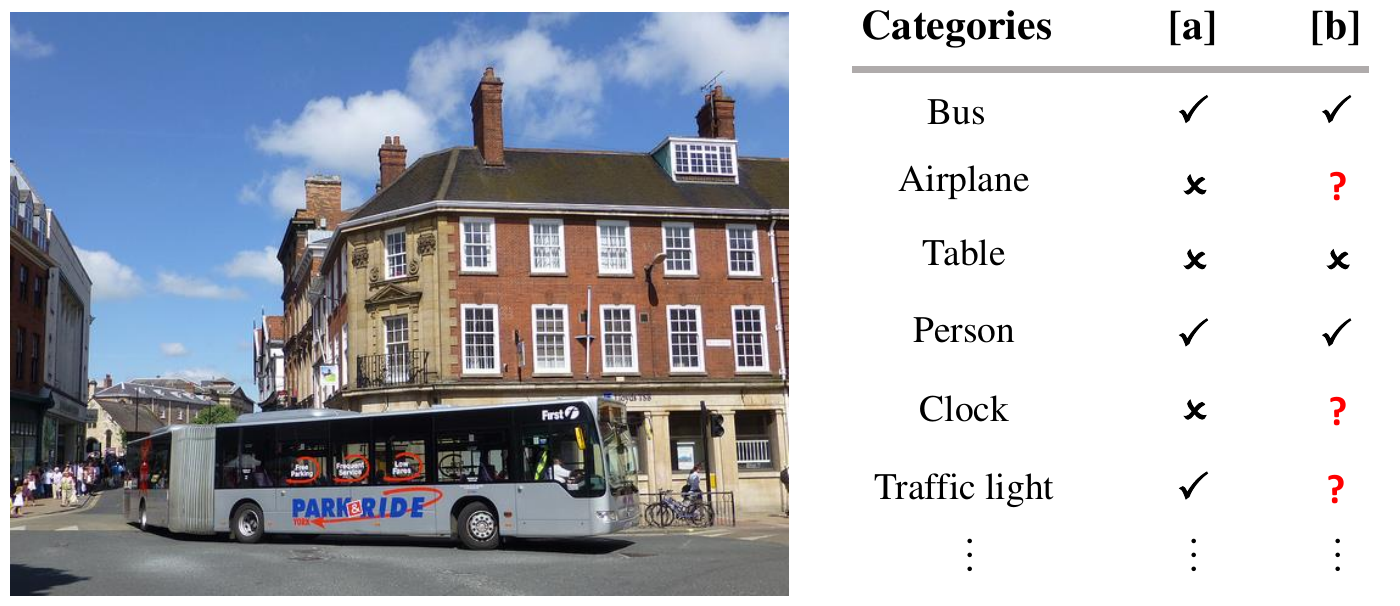}
	\caption{An illustration of MLR image with complete labels [a] and partial labels [b]. All positive and negative labels are known in traditional MLR, while some labels are missing in MLR-PL (airplane, clock, traffic light).}
    \Description{A figure containing a bus and other things, besides with its annotations.}
	\label{Fig:Setting}
\end{figure*}

Multi-label image recognition (MLR) is a fundamental yet practical task in computer vision. Since images in the real-world generally contain multiple instances of different categories, MLR provides a more comprehensive understanding in vision systems compared to single-label image recognition. It also benefits various applications ranging from content-based image retrieval \cite{cheng2005semantic, li2010technique, zhang2021instance, lai2016instance}, recommendation systems \cite{carrillo2013multi, darban2022ghrs, zheng2014context} to human attribute analysis \cite{chen2021cross, pu2021expression, wu2019instance}. Many significant progress has been made in this task thanks to the development of deep neural networks in recent years \cite{he2016resnet, dosovitskiy2020vit}. However, these methods rely on large-scale and high quality datasets with clean and complete annotations for training, while it is very time-consuming and labor-intensive to annotate a consistent and complete list of labels for each image in multi-label task. With this consideration, we consider the task of multi-label recognition with partial labels (MLR-PL), where only part of labels are known in each image, while others are unknown, as shown in Figure \ref{Fig:Setting}. Such annotation strategy is more flexible and scalable as the number of samples and categories are always expanding in real application scenarios.

Current algorithms for the MLR-PL task mainly exploit label dependencies or feature similarity to transfer knowledge from known labels to unknown labels and gain extra supervision information for model training \cite{wang2021pico, chen2022structured, pu2022semantic}. Despite significant progress, these methods still rely on substantial amounts of data to learn category relations and discriminative features. Additionally, they require complex framework designs to obtain additional supervisory information during training, which demands intricate training strategies and dataset-specific tuning of hyperparameters. These factors limit the scalability and performance in practical applications. Recently, vision-language models (VLM) such as CLIP \cite{radford2021CLIP} and ALIGN \cite{jia2021scaling} have demonstrated great potential in label-limited tasks, benefiting from rich visual knowledge acquired through training on large-scale datasets with image-text pairs. Therefore, we propose leveraging the prior knowledge of CLIP to establish a framework for MLR-PL without the need for extensive training data. 
Prompt learning provides an efficient way to adapt VLM to other tasks, which is implemented by setting manual templated prompts or learnable prompt tokens to avoid costs for fine-tuning the entire model. Through this method, current works like CoOp \cite{zhou2022coop} have demonstrated their effectiveness and generalization ability to various few shot visual tasks. 
However, these methods mainly extract a global representation of the entire image using the visual encoder and utilize a single prompt text for all categories, which is suitable for single-label classification with only one dominant category. In contrast, in MLR-PL, images contain multiple semantic objects, leading to entangled category information in the global representation. Moreover, the semantics of the global representation tend to be dominated by salient objects after coarse-grained operations (e.g., pooling). These factors result in semantic confusion when matching with text features, particularly when features across different categories share similarities, thereby diminishing the model’s performance. This enables us to explore methods for extracting fine-grained visual semantic information from CLIP's global visual representation, which is essential for MLR-PL.

Building on the analysis above, we enhance the original CLIP's capability in MLR-PL by integrating a fine-grained approach to obtain decoupled representations for each category. Specifically, we introduce a semantic-guided spatial attention mechanism on top of the CLIP visual encoder, which locates informative regions for different semantic objects and separates category-specific feature maps from the semantically entangled global representation. The existence of a target class is then determined by matching the decoupled visual features with textual representations, which effectively alleviates the semantic confusion arising from semantic uncertainty. Moreover, considering the diversity of features across different categories in MLR, adopting a unified prompt template is limited in expressiveness and lacks category-specific descriptive information. It is also challenging for the model to integrate the semantics of all categories into a single prompt template, which restricts its performance in MLR-PL. To address these problems, we introduce a category-specific prompt optimization method to learn text representations that are aligned with category semantics, further decoupling semantic information across categories.

Our main contributions in this paper are summarized as follows.
\begin{itemize}
\item We introduce a simple yet effective framework for MLR-PL task based on vision-language models. To alleviate the semantic confusion caused by entangled global representation, we incorporate a semantic decoupling module based on a semantic-guided spatial attention mechanism to learn semantic-specific features. We also utilize category-specific prompt optimization method for learning the prompts that best aligned with category semantics. During training, only the semantic-decoupling module and prompt optimization module are learnable, while the remaining parameters of the vision-language model are kept fixed. This approach avoids fine-tuning the entire model. 
\item We conduct extensive experiments on various datasets (e.g., Microsoft COCO 2014 and Pascal VOC 2007), which demonstrate superior performance compared to traditional leading methods and previous CLIP-based approaches. Additionally, we conduct ablation studies to analyze the contribution of semantic-decoupling module and provide visualization analysis on our method for a more comprehensive understanding.
\end{itemize}

\section{Related Works}
\label{sect:related}
\subsection{Multi-label Recognition}

Multi-label recognition has drawn increasing attention in recent years \cite{Zhou2023Algin, Zhou2023AttentionAug, chen2024dynamic}. One straightforward method for this task is to view it as a multiple-binary classification task. This method ignores the label correlations among categories, which is crucial in multi-label recognition task. Hence, recent works proposed exploiting the underlying label structure to regularize the training of MLR models. One way to achieve this goal is by introducing recurrent neural network (RNN)/long short-term memory (LSTM) \cite{hochreiter1997long,wang2017recurrent} to capture label dependencies implicitly. Wang et al. \cite{wang2016cnn-rnn} proposed a CNN-RNN framework that model label dependency with semantic redundancy and co-occurrence dependency. Additionally, Chen et al. \cite{chen2018recurrent} proposed a recurrent attention reinforcement learning framework to model long-term dependencies among  attentional regions and capture semantic label co-occurrences. Another line of works explicitly model label dependencies with graph neural network (GNN) \cite{li2015gated, kipf2016semi, Chen2023GraphAttention, Zhou2023DoubleAttention}. For example, Chen et al. \cite{chen2019multi} introduced an inter-dependent object classifier built on a graph convolutional neural network (GCN), which is trained to model the correlations between different categories. Moreover, Chen et al. \cite{chen2019learning} proposed a semantic-specific graph representation learning (SSGRL) framework that utilizes a graph propagation scheme to obtain semantic-aware features, which achieved state-of-the-art performance on several multi-label datasets. Despite achieving remarkable performance in various datasets and application scenarios, these methods are built on deep neural network (DNN) that require large-scale and complete annotated datasets for training. However, it is very time-consuming and labor-intensive to annotate a complete label list for each image, which makes the collection of these datasets less practical and thus limits the application of MLR.

\subsection{Multi-label Recognition With partial labels}
To reduce the annotation cost, many works have been dedicated to the task of multi-label recognition with partial labels (MLR-PL), where only part of labels are known in each image. Current works tend to introduce pseudo-labels for the complement of unknown labels. For example, Durand et al. \cite{durand2019learning} introduced the curriculum learning strategy where pseudo-labels are generated by the pre-trained model, and used the partial-BCE loss normalized by the proportion of known labels for model training. Moreover, Huynh et al. \cite{huynh2020interactive} proposed to regularize the BCE loss with the statistical co-occurrences and image-level feature similarity. Recently, some works employ category-specific feature blending \cite{pu2022semantic, pu2024dual} or label correlations between categories \cite{chen2022structured, Chen2024HST} to transfer information from known labels and supplement extra pseudo-labels for model training. These methods basically rely on the modeling of label dependencies. Other works propose reject or correct samples with large loss fluctuation to prevent the model from memorizing the noisy labels \cite{kim2022large}, or enhance the salient object regions corresponding to the present labels \cite{wang2023saliency, kim2023bridging}.

Although achieving significant progress, these MLR-PL methods rely on complex model architectures and training strategies. Motivated by recent progress in vision-language model (VLM), our approach suggests to transfer knowledge from large-scale pre-trained VLM by prompt optimization and thus provides a simple framework for solving the MLR-PL task.

\subsection{Vision-Language Models}
Vision-language pre-training models \cite{jia2021scaling, radford2021CLIP} have demonstrated impressive performance on various downstream tasks. Among these models, CLIP \cite{radford2021CLIP} is a representative framework which leverage large-scale image-text pairs collected from the Internet to align image and text representations in the embedding space. It has obtained rich visual representations and knowledge through large-scale contrastive learning and thus demonstrated remarkable performance and scalability in zero-shot inference and few-shot tasks\cite{du2022openvoca, He2023openvoca, kim2023regionopenvoca, qin2023freesegopenvoca}. To fit models to downstream tasks while preserving the learned representation space, recent approaches mainly adopt prompt-tuning method\cite{DAprompt, lu2022prompt, bulat2023lasprompt, vlmprompt} instead of fine-tuning the entire model. It provides a parameter-efficient and flexible way and is ease of use. In CoOp \cite{zhou2022coop}, Zhou et al.proposed to learn the prompts from the target dataset and avoid manually fine-tuning prompt templates. CoCoOp \cite{zhou2022cocoop} further enhanced CoOp by introducing image-conditional information to improve generalization to unseen classes. However, although achieving remarkable process, CLIP is trained to focus on global image and text representations. Consequently, it tends to predict the closest semantic class while neglecting other potential classes during inference. Since images contain multiple objects in MLR task, fine-grained information is essential for the prediction accuracy. For similar consideration, RegionCLIP \cite{zhong2022regionclip} proposed to learn region-level visual features in object detection and align the region-text pairs to make prediction. However, large-scale dataset for object detection is needed to fine-tune the detector, which does not match the need of partial label learning. Different from the above approaches, our method proposes to learn semantic-specific representations to better disentangle visual and textual information from different categories to avoid semantic confusion caused by global representation scheme.

\section{Proposed Method}

\begin{table}[h]
    \centering
    \caption{Mathematical Notations and Descriptions}
    \begin{tabular}{c c}
    \hline
    \textbf{Notation} & \textbf{Description} \\ \hline
    $\mathcal{D}$ & The training dataset \\ 
    $N$ & The number of samples in training dataset \\ 
    $I^n$ & The $n$-th image in the dataset \\ 
    $y^n$ & The label vector for the $n$-th image \\ 
    $Enc_v$ & The visual encoder in CLIP with slight modification \\
    $Enc_t$ & The textual encoder in CLIP \\
    $F_{att}$ & The attention mapping function in semantic decoupling module \\
    $F_g$ & The pre-trained language model GloVe \cite{pennington2014glove} \\
    $\odot$ & The Hadamard product between features \\
    $\widetilde{a}_{c, wh}$ & The semantic attention coefficient for category $c$ at position $(w, h)$ \\ 
    $a_{c, wh}$ & The normalized attention coefficient for category $c$ at position $(w, h)$ \\
    $f^g_{wh}$ & The global visual representation from CLIP visual encoder at position $(w, h)$ \\
    $\widetilde{f}_{c, wh}$ & Spatial feature fused with semantic information for category $c$ at $(w, h)$ \\
    $f^v_c$ & Visual feature for category $c$ \\
    $f^t_c$ & Textual feature for category $c$ \\
    $x_c$ & The semantic representation of category $c$  \\
    $t_c$ & The learnable prompt for category $c$ \\
    $[V]_c^i$ & The $i$-th learnable word vector in $t_c$ \\
    $[CLS]$ & The word embedding of the category name \\
    $p_c$ & Model's prediction score for category $c$ \\
    $\tau$ & The temperature parameter in softmax \\
    $\gamma_{+}$ & The exponential coefficient in P-ASL for positive samples \\
    $\gamma_{-}$ & The exponential coefficient in P-ASL for negative samples \\
    $\mathcal{L}^n_c$ & The loss term for category $c$ of the $n$-th sample \\
    $\mathcal{L}$ & The overall loss function \\ \hline
    \end{tabular}
    \label{notaion}
\end{table}

We formulate the MLR-PL task as follows. Let $\mathcal{D} = \{(I^1, y^1), ..., (I^N, y^N)\}$ be the training set containing $N$ training samples and $C$ categories, where $I^n$ denotes the $n$-th image and $y^n=[y^n_1,...,y^n_C] \in \{-1,0,1 \}^C$ denotes its label vector. Each image may contain objects of multiple categories. In partial label setting, $y^n_c$ is assigned to 1 if the category $c$ is present in the $n$-th image, assigned to -1 if it is absent, or assigned to 0 if it is unknown. 

\begin{figure*}[!t] 
	\centering 
	\includegraphics[width=0.9\linewidth]{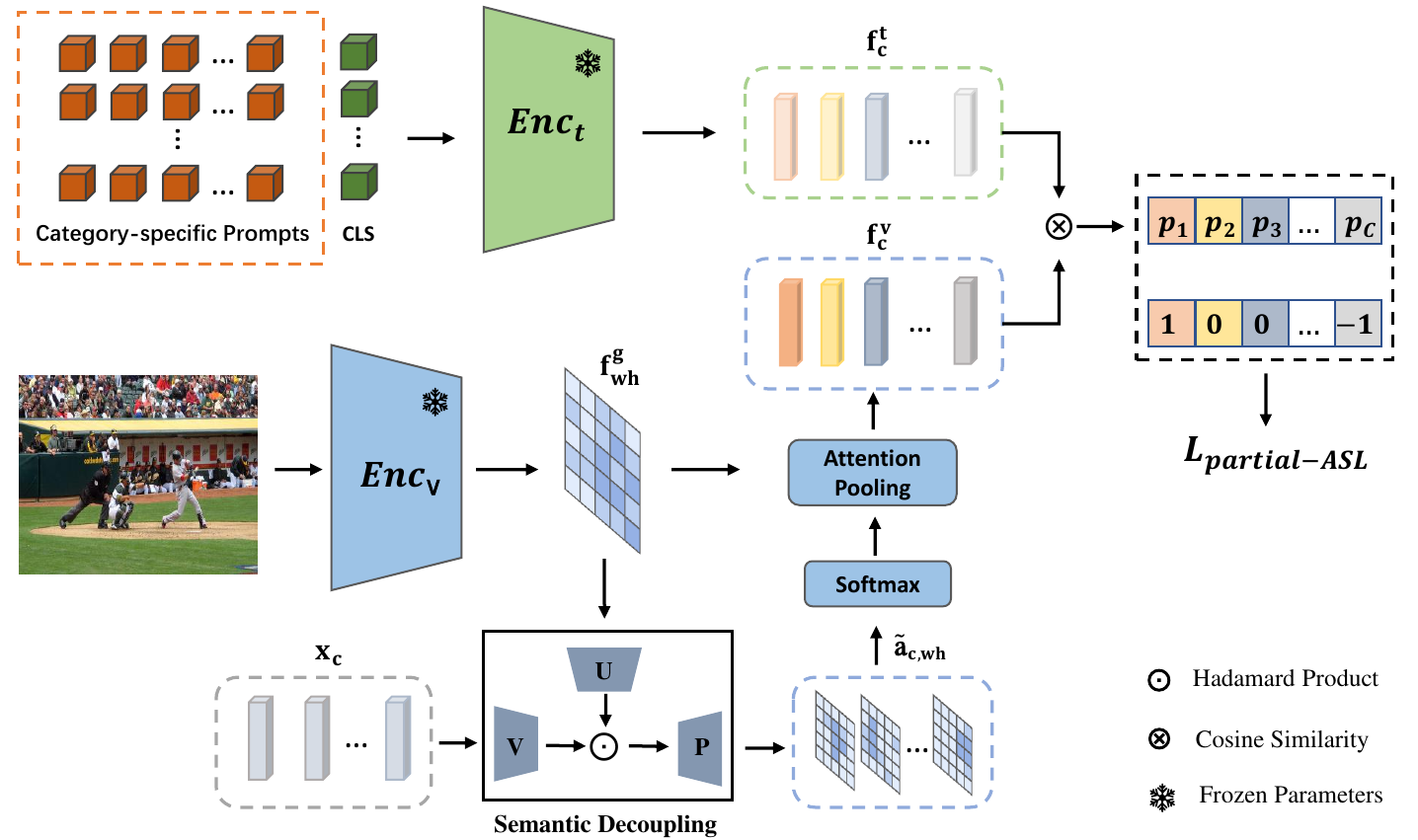}
	\caption{An overall illustration of the proposed framework. For each class, a prompt is initialized with each component as a learnable vector. These prompts are passed through CLIP textual encoder to obtain category-specific text embeddings. Meanwhile, the semantic decoupling module employs a semantic-guided spatial attention mechanism to extract fine-grained visual features for each category. The parameters of semantic decoupling module and the category-specific learnable prompts are optimized by minimizing the classification loss, while the parameters of CLIP visual and textual encoders remain fixed.}
    \Description{Pipeline for CLIP-based MLR-PL framework.}
	\label{Fig:Fram}
\end{figure*}

In this section, we introduce the proposed framework which builds on pre-trained VLM with rich prior information. To alleviate the semantic confusion problem caused by utilizing single global visual and textual representation for all categories in current VLM methods, the visual semantic decoupling module first extracts category-specific features from entangled global representation. It employs a semantic-guided spatial attention mechanism to locate informative regions for specific category, and obtains visual features that encode the category-specific information. Furthermore, we set category-specific prompts that best align with the semantics of certain category. To avoid manual tuning, we propose using prompt optimization method to update prompt tokens through the training process. 
The prediction probability for each category is calculated by coupling the category-specific visual feature with the corresponding textual feature. 
Finally, the model supervises the parameter learning of the semantic decoupling module and the category-specific prompts by minimizing the classification loss, while the parameters of visual and textual encoders remain fixed during training. We adopt the partial asymmetric loss widely used in MLR, in which only the annotated samples are considered for calculating the loss. An overall illustration of our framework is shown in Figure \ref{Fig:Fram}. We also include the notations used in this section in Table \ref{notaion} for clarity.

\subsection{Visual Semantic Decoupling}
The semantic decoupling module is introduced to extract category-specific information from the global feature maps of semantic entanglement. It is implemented by introducing a semantic-guided spatial attention mechanism. Specifically, given an image $I$, the framework first utilizes the visual encoder in CLIP to extract its global feature map $f^g  \in \mathbb{R}^{N \times W \times H}$ as follows, where $N$, $W$ and $H$ are the channel numbers, width and height of the feature map,
\begin{equation}\label{eq1}
    f^g=Enc_v(I),
\end{equation}
and $Enc_v$ is the visual encoder in CLIP with slight modification. To extract semantic information of certain category $c$, we utilize pre-trained language model GloVe \cite{pennington2014glove} to obtain the semantic representation $x_c \in \mathbb{R}^{D}$ as guidance, formulated as
\begin{equation}\label{eq2}
    x_c=F_{g}(w_c),
\end{equation}
where $w_c$ is the semantic word of category $c$ and $F_{g}$ is the pre-trained GloVe model. We can then incorporate $f^g$ with $x_c$ to focus on spatial regions with semantic information of category $c$. Specifically, for each location $(w, h)$, we fuse these information with a low-rank bilinear pooling method \cite{kim2016hadamard}:
\begin{equation}\label{eq3}
    \widetilde{f}_{c,wh}=P^T(tanh((U^T f^g_{wh}) \odot (V^T x_c)))+b,
\end{equation}
where $f^g_{wh}$ is the local feature of $f^g$ at position $(w, h)$, $P, U, V$ are learnable linear functions, $b$ is a learnable bias vector, and $\odot$ represents the Hadamard product. $\widetilde{f}_{c,wh}$ is the spatial feature that fuses category semantic. We then utilize an attention mapping function $F_{att}$ implemented by a fully connected layer to compute the spatial attention map as follows.
\begin{equation}\label{eq4}
    \widetilde{a}_{c, wh}=F_{att}(\widetilde{f}_{c, wh}).
\end{equation}
The coefficient at position $(w,h)$ on the map represents the semantic relevance of the region to category c. Additionally, to make these attention coefficients more comparable across different samples, we normalize them with a softmax function over all regions on the map, formulated as
\begin{equation}\label{eq5}
    a_{c, wh}=\frac{exp(\widetilde{a}_{c, wh})}{\sum_{w',h'}exp(\widetilde{a}_{c, w'h'})}.
\end{equation}

Finally, we extract the spatial-related representation $f^v_c$ by applying attention pooling for the $c$-th category as follows:
\begin{equation}\label{eq6}
    f_c^v=\sum_{w,h} a_{c, wh} \odot f^g_{c, wh}.
\end{equation}
The $f_c^v$ encodes all semantic information related to category $c$. Consequently, the semantic decoupling module disentangles the global representation into distinct semantic components. By repeating this procedure, the model obtains category-specific representations $\{f_1^v, f_2^v,..., f_C^v \}$ for all categories. We further present visualization illustrations to demonstrate the effectiveness of the module, as shown in Figure \ref{Fig:CAM}.

\subsection{Category-specific Prompt Optimization}
Due to the fine-grained characteristics of MLR task, we propose setting category-specific prompts that best align with the given semantic contexts. To avoid manual prompt tuning, we follow CoOp \cite{zhou2022coop} by modeling the tokens as continues, learnable parameters that can be optimized in an end-to-end manner with classification loss, while keeping the parameters of pre-trained text encoder fixed. Specifically, we define the specific prompt $t_c$ for category $c$ as follows: 
\begin{equation}\label{eq7}
    t_c=[V]_c^1 [V]_c^2 ... [V]_c^M [CLS],
\end{equation}
where $M$ is the length of prompt. Each $[V]^i_c$ is a learnable word vector (with a dimension of 512 in CLIP) and $[CLS]$ represents the word embedding corresponding to the category name. These tokens are randomly initialized by sampling from a Gaussian distribution and are independent for each category. we obtain the textual representation for the $c$-th category as follows:
\begin{equation}\label{eq8}
    f_c^t = Enc_t(t_c),
\end{equation}
where $Enc_t$ is the text encoder of CLIP. By coupling the visual and textual features from each category, the prediction score for category $c$ is computed as:
\begin{equation}\label{eq9}
    p_c=\frac{exp ( \langle f_c^v, f_c^t \rangle / \tau )}{\sum_{c=1}^{C}exp ( \langle f_c^v, f_c^t \rangle / \tau )},
\end{equation}
where $\langle \cdot, \cdot \rangle$ calculate the cosine similarity between two features, $c \in \{1, 2, ..., {C}\}$ is the category index, and $\tau$ is the temperature parameter.

\subsection{Model Optimization}
During the training process, only the semantic decoupling module and category-specific prompts are updated, while  the parameters of visual and textual encoders remain fixed. The optimization process is performed to minimize the classification loss. To address the inherent positive-negative imbalance in MLR datasets, we adopt partial Asymmetric Loss (P-ASL) \cite{ridnik2021asymmetric} to alleviate this problem. Specifically, given the model prediction scores $p^n=[p^n_1, ..., p^n_C]$ for the $n$-th sample in training set, only the annotated samples are utilized to calculate the loss, which is formulated as
\begin{equation}\label{eq:10}
    \mathcal{L}^n_c=
    \begin{cases}
        (1-p^n_c)^{\gamma_{+}} log(p^n_c) & y^n_c=1 \\
        (\Bar{p}^n_{c})^{\gamma_{-}} log(1-\Bar{p}^n_c) & y^n_c=-1 ,\\
    \end{cases}
\end{equation}
where $\Bar{p}_c^n=max(p_c^n-m, 0)$ represents the probability for negative samples shifted by a hard threshold $m$ which is set to down-weight the easy negative samples. We also set the focusing parameter $\gamma_{-} \ge \gamma_{+}$ to further reduce the contribution of easy negative samples. The semantic decoupling module and learnable prompts are updated through back-propagation using P-ASL. Finally, the objective for training is formulated as follows:
\begin{equation}\label{eq:11}
    \mathcal{L} = -\frac{1}{N} \sum_{n=1}^N \sum_{c=1}^C \mathcal{L}_c^n.
\end{equation}

The training procedure of our proposed framework is summarized in Algorithm \ref{alg:Framwork}.

\renewcommand{\algorithmicrequire}{ \textbf{Input:}}     
\renewcommand{\algorithmicensure}{ \textbf{Output:}}    

\begin{algorithm}[!ht]
\caption{Training procedure of the proposed framework.}
\label{alg:Framwork}
\begin{algorithmic}[1] 
    \REQUIRE $\mathcal{D}=\{(I^n,y^n)\}_{n=1}^{N}$: training dataset with partial labels \\
    \quad \quad $Enc_v, Enc_t$: visual and textual encoders from CLIP \\
    \quad \quad $\{t_1, t_2, ..., t_C \}, \mathcal{F}_{sd}$: initialized category-specific prompts and semantic decoupling module. \\
    \ENSURE updated parameters of $\{t_1, t_2, ..., t_C \}$ and $\mathcal{F}_{sd}$\\

    \STATE Initialize parameters of $\{t_1, t_2, ..., t_C \}$ and $\mathcal{F}_{sd}$ randomly;
    \FOR {$epoch =1, 2, ..., E_{max}$}
        \STATE Extract the category-specific visual features: $\{f_1^v, f_2^v, ..., f_C^v \}=\mathcal{F}_{sd}(Enc_v(I))$;
        \STATE Extract category-specific text embeddings: $\{f_1^t, f_2^t, ..., f_C^t\}=Enc_t(t_1, t_2, ..., t_C)$;
        \STATE Compute categorical prediction scores $\{p_1,p_2,...,p_C \}$ with Equation \ref{eq9};
        \STATE Compute the value of partial asymmetric loss with Equation \ref{eq:11};
        \STATE Update the parameters of $\{t_1, t_2, ..., t_C \}$ and $\mathcal{F}_{sd}$ with SGD;
    \ENDFOR
\end{algorithmic}
\end{algorithm}

\section{Experiments}

In this section, we evaluate the performance of our framework through extensive experiments and comparisons with current leading MLR-PL methods. Furthermore, we present visual illustrations to demonstrate the effectiveness of the proposed modules.

\subsection{Datasets and Evaluation Metrics}
We conduct experiments and comparisons on the MS-COCO \cite{lin2014coco} and Pascal VOC 2007 \cite{everingham2010voc} datasets, which are the most widely used benchmarks in MLR. The MS-COCO dataset contains about 120k images and 80 different categories in daily life. The dataset is further divided into a training set with around 80k images and a testing dataset with about 40k images. The Pascal VOC 2007 dataset contains around 10k images with 20 object categories, and it is divided into a training set and a validation set with 5,011 images and 4,952 images, respectively.
Since both of these datasets are completely-annotated, we randomly drop certain proportions of labels to create training datasets with partial-labels. In our experiments, the proportions of known labels in datasets vary from 10\% to 90\%. The testing datasets retain complete labels to ensure adequate samples for evaluation.

For the evaluation metrics, we follow current works and mainly adopt the mean average precision (mAP) over all categories under different proportions of known labels. Additionally, we compute the average mAP across all proportions for a more comprehensive comparison. Furthermore, we follow current MLR works to adopt the overall and per-class F1-measure (i.e., the OF1 and CF1 metrics), which are defined as follows:
\begin{equation}
    OP=\frac{\sum_i N^c_i}{\sum_i N^p_i}, \quad CP=\frac{1}{C} \frac{\sum_i N^c_i}{\sum_i N^p_i},
\end{equation}
\begin{equation}
    OR=\frac{\sum_i N^c_i}{\sum_i N^g_i}, \quad CR=\frac{1}{C} \frac{\sum_i N^c_i}{\sum_i N^g_i},
\end{equation}
\begin{equation}
    OF1=\frac{2 \times OP \times OR}{OP+OR}, \quad CF1=\frac{2 \times CP \times CR}{CP+CR},
\end{equation}
where $C$ is the number of labels, $N^c_i$ is the number of images that are correctly predicted in the $i$-th category, $N^p_i$ is the number of predicted in the $i$-th category, and $N^g_i$ is the number of ground truth images in the $i$-th category. We also compute the average OF1 and CF1 scores across all proportions of known labels.

\subsection{Implementation Details}
We adopt ResNet-101 \cite{he2016resnet} from the CLIP pre-trained model as the visual encoder to extract global representation, and the input size is 448$\times$448. We replace the last average pooling layer in ResNet with another average pooling layer with a size of 2$\times$2 and a stride of 2 while keeping other layers unchanged. We also use the same Transformer in CLIP as the text encoder. Both of the visual and text encoder are frozen during training. For visual semantic decoupling module, the attention mapping function is implemented by a fully connected layer that maps vector from dimension 1024 to 1. We also set an intermediate layer implemented by a 1024-to-512 linear map to cope with the dimension of text features in each category. For category-specific prompts, we set independent prompt vector with 16 tokens (M=16) for each category. During training process, only the category-specific prompts and the visual semantic decoupling module are trainable. The total number of trainable parameters is 4.8M.

The proposed framework is trained by minimizing the loss $\mathcal{L}$ as shown in Equation \ref{eq:11}. In partial Asymmetric Loss, we set $\gamma_{-}=2$, $\gamma_{+}=1$ to better down-weight easy negative samples, and the hard threshold $m$ for negative probability is set to 0.05. We use the SGD optimizer with initial learning rate of 0.002, which is decayed by the cosine annealing rule during training. We also use warm-up strategy with learning rate of 0.0005 in the first epoch. The model is trained with batch-size of 64 and epoch of 100. Our model is implemented in PyTorch \cite{pytorch2019}, and training is conducted on one 24GB NVIDIA RTX 3090 GPU.

\begin{table*}[!t] 
    \centering 
    \tiny
    \setlength{\tabcolsep}{2pt}
    \caption{The average mAP and mAP values achieved by our framework and current state-of-the-art methods for MLR-PL under different known label proportions on the MS-COCO and Pascal VOC 2007 datasets. The best results are highlighted in bold. The proposed method achieves superior performance compared to current mainstream methods under all settings of known labels proportions.}
    \vspace{5pt}
    \scalebox{1.0}{
    \begin{tabular*}{\textwidth}{@{\extracolsep{\fill}} c|c|ccccccccc|c}
    \hline
    Datasets &Methods &10\% &20\% &30\% &40\% &50\% &60\% &70\% &80\% &90\% &Avg.\\ 
    \hline \noalign{\smallskip} \hline
    \multirow{8}{*}{\centering MS-COCO}
    &SSGRL \cite{chen2019learning}  &$62.4\pm0.4$ &$70.5\pm0.4$ &$73.2\pm0.3$ &$74.6\pm0.2$ &$76.3\pm0.2$ &$76.5\pm0.1$ &$77.1\pm0.1$ &$77.9\pm0.1$ &$78.4\pm0.1$ &$74.1\pm0.2$\\
    &GCN-ML \cite{chen2019multi}  &$63.8\pm0.2$ &$70.9\pm0.2$ &$72.8\pm0.2$ &$74.0\pm0.2$ &$76.7\pm0.2$ &$77.1\pm0.2$ &$77.3\pm0.1$ &$78.3\pm0.1$ &$78.6\pm0.1$ &$74.4\pm0.2$\\
    &KGGR \cite{chen2022knowledge}  &$66.6\pm0.1$ &$71.4\pm0.2$ &$73.8\pm0.1$ &$76.7\pm0.1$ &$77.5\pm0.1$ &$77.9\pm0.1$ &$78.4\pm0.1$ &$78.7\pm0.1$ & $79.1\pm0.1$ &$75.6\pm0.1$\\
    &SST \cite{chen2022structured}  &$68.1\pm0.3$ &$73.5\pm0.2$ &$75.9\pm0.2$ &$77.3\pm0.2$ &$78.1\pm0.1$ &$78.9\pm0.1$ &$79.2\pm0.1$ &$79.6\pm0.1$ &$79.9\pm0.1$ &$76.7\pm0.1$\\
    &HST \cite{Chen2024HST}  &$70.6\pm0.4$ &$75.8\pm0.3$ &$77.3\pm0.3$ &$78.3\pm0.2$ &$79.0\pm0.2$ &$79.4\pm0.1$ &$79.9\pm0.1$ &$80.2\pm0.1$ &$80.4\pm0.1$ &$77.9\pm0.2$\\
    &SARB \cite{pu2022semantic} &$71.2\pm0.3$ &$75.0\pm0.3$ &$77.1\pm0.3$ &$78.3\pm0.3$ &$78.9\pm0.3$ &$79.6\pm0.2$ &$79.8\pm0.2$ &$80.5\pm0.2$ &$80.5\pm0.2$ &$77.9\pm0.3$\\
    &DSRB \cite{pu2024dual} &$72.5\pm0.6$ &$76.0\pm0.5$ &$77.6\pm0.5$ &$78.7\pm0.4$ &$79.6\pm0.4$ &$79.8\pm0.3$ &$80.0\pm0.2$ &$80.5\pm0.2$ &$80.8\pm0.2$ &$78.4\pm0.3$\\
    &\textbf{Ours} &\textbf{78.3} $\pm$ 0.2 &\textbf{80.2} $\pm$ 0.2 &\textbf{81.2} $\pm$ 0.2 &\textbf{81.6} $\pm$ 0.2 &\textbf{82.3} $\pm$ 0.1 &\textbf{82.4} $\pm$ 0.1 &\textbf{82.6} $\pm$ 0.1 &\textbf{83.4} $\pm$ 0.1 &\textbf{83.6} $\pm$ 0.1 &\textbf{81.7} $\pm$ 0.2\\
    \hline \noalign{\smallskip} \hline
    \multirow{8}{*}{\centering Pascal VOC}
    &SSGRL \cite{chen2019learning}  &$77.7\pm0.9$ &$87.6\pm0.6$ &$89.9\pm0.4$ &$90.7\pm0.3$ &$91.4\pm0.3$ &$91.8\pm0.3$ &$92.0\pm0.2$ &$92.2\pm0.2$ &$92.2\pm0.2$ &89.5 $\pm$ 0.4\\
    &GCN-ML \cite{chen2019multi} &$74.5\pm0.3$ &$87.4\pm0.2$ &$89.7\pm0.2$ &$90.7\pm0.2$ &$91.0\pm0.2$ &$91.3\pm0.2$ &$91.5\pm0.2$ &$91.8\pm0.1$ &$92.0\pm0.1$ &88.9 $\pm$ 0.2\\
    &KGGR \cite{chen2022knowledge}  &$81.3\pm0.1$ &$88.1\pm0.2$ &$89.9\pm0.1$ &$90.4\pm0.1$ &$91.2\pm0.1$ &$91.3\pm0.1$ &$91.5\pm0.1$ &$91.6\pm0.1$ &$91.8\pm0.1$ &89.7 $\pm$ 0.1\\
    &SST \cite{chen2022structured} &$81.5\pm0.2$ &$89.0\pm0.2$ &$90.3\pm0.2$ &$91.0\pm0.2$ &$91.6\pm0.1$ &$92.0\pm0.1$ &$92.5\pm0.1$ &$92.6\pm0.1$ &$92.7\pm0.1$ &90.4 $\pm$ 0.2\\
    &HST \cite{Chen2024HST} &$84.3\pm0.4$ &$89.1\pm0.3$ &$90.5\pm0.3$ &$91.6\pm0.2$ &$92.1\pm0.2$ &$92.4\pm0.1$ &$92.5\pm0.1$ &$92.8\pm0.1$ &$92.8\pm0.1$ &90.9 $\pm$ 0.2\\
    &SARB \cite{pu2022semantic} &$83.5\pm0.3$ &$88.6\pm0.3$ &$90.7\pm0.3$ &$91.4\pm0.3$ &$91.9\pm0.3$ &$92.2\pm0.3$ &$92.6\pm0.2$ &$92.8\pm0.2$ &$92.9\pm0.2$ &90.7 $\pm$ 0.3\\
    &DSRB \cite{pu2024dual} &$85.7\pm0.6$ &$89.8\pm0.6$ &$91.8\pm0.5$ &$92.0\pm0.4$ &$92.3\pm0.4$ &$92.7\pm0.3$ &$92.9\pm0.2$ &$93.1\pm0.2$ &$93.2\pm0.2$ &91.5 $\pm$ 0.3\\
    &\textbf{Ours} &\textbf{88.5} $\pm$ 0.2 &\textbf{91.5} $\pm$ 0.2 &\textbf{92.6} $\pm$ 0.2 &\textbf{93.2} $\pm$ 0.2 &\textbf{93.3} $\pm$ 0.1 &\textbf{93.5} $\pm$ 0.1 &\textbf{93.8} $\pm$ 0.1 &\textbf{94.1} $\pm$ 0.1 &\textbf{94.3} $\pm$ 0.1 &\textbf{92.8} $\pm$ 0.2\\
    \hline
\end{tabular*}
}
\label{table1} 
\end{table*}

\begin{table*}[!t]
    \centering
    \caption{The average OF1 and CF1 values achieved by our framework compared to current state-of-the-art methods for MLR-PL on the MS-COCO and Pascal VOC 2007 datasets.}
    \begin{tabular}{c|c|c|c}
    \hline
        Datasets & Methods &Avg.OF1 &Avg.CF1\\
        \hline \noalign{\smallskip} \hline
        \multirow{8}{*}{\centering MS-COCO}
         & SSGRL \cite{chen2019learning} &73.9 &68.1  \\
         & GCN-ML \cite{chen2019multi} &73.1 &68.4  \\
         & KGGR \cite{chen2022knowledge}  &73.7 &69.7 \\
         & SST \cite{chen2022structured} &75.8 &71.2 \\
         & HST \cite{Chen2024HST} &76.7 &72.6 \\
         & SARB \cite{pu2022semantic} &76.5 &72.2 \\
         & DSRB \cite{pu2024dual} &76.8 &72.7\\
         & \textbf{Ours} &\textbf{77.9} &\textbf{75.1} \\
        \hline \noalign{\smallskip} \hline
        \multirow{8}{*}{\centering Pascal VOC}
         & SSGRL \cite{chen2019learning} &87.7 &84.5  \\
         & GCN-ML \cite{chen2019multi} &87.3 &84.6  \\
         & KGGR \cite{chen2022knowledge} &86.5 &84.7 \\
         & SST \cite{chen2022structured} &88.2 &85.6 \\
         & HST \cite{Chen2024HST} &88.4 & 86.1 \\
         & SARB \cite{pu2022semantic} &88.4 &85.9 \\
         & DSRB \cite{pu2024dual} &88.3 &86.0\\
         & \textbf{Ours} &\textbf{88.6} &\textbf{86.5} \\
        \hline
    \end{tabular}
    \label{table2}
\end{table*}

\subsection{ Comparison with State-of-the-Art Methods}
To evaluate the effectiveness of the proposed framework, we compare it with the following baseline methods, which can be divided into three groups. (1) SSGRL (ICCV'19) \cite{chen2019learning}, GCN-ML (CVPR'19) \cite{chen2019multi}, and KGGR (TPAMI'22) \cite{chen2022knowledge}. These methods adopt graph neural networks to model label dependencies, and they achieved leading performance in MLR with complete labels. In MLR-PL task, we adapt these methods by replacing the BCE loss with partial BCE loss \cite{durand2019learning} during training. (2) SST (AAAI'22) \cite{chen2022structured} and HST (IJCV'24) \cite{Chen2024HST}. These methods generate pseudo-labels by exploring semantic correlations of visual features within and across images. HST further proposed a dynamical threshold learning method to  adaptively search the optimal threshold for pseudo-label generation. (3) SARB (AAAI'22) \cite{pu2022semantic} and DSRB (ESWA'24) \cite{pu2024dual}. These methods blend category-specific representation across different images to complement unknown labels with information from known labels.

For the MS-COCO dataset, Table \ref{table1} and \ref{table2} shows the comparisons of mAP and average OF1 and CF1 scores with all baseline methods under known label proportions of 10\% to 90\%. Our framework significantly outperforms other baseline methods under all proportions of known labels, particularly when the label proportions are extremely low. Specifically, it obtains mAPs of 78.3\%, 80.2\%, 81.2\%, 81.6\%, 82.3\%, 82.4\%, 82.6\%, 83.4\% and 83.6\% under the settings of 10\%-90\% known labels, respectively. Additionally, our framework obtains average mAP, OF1 and CF1 values of 81.7\%, 77.9\%, 75.1\%, with average improvements over the second-best method by 3.3\%, 1.1\%, 2.4\% respectively. 

For the Pascal VOC 2007 dataset, we also present the comparison results in Table \ref{table1} and \ref{table2}. Since the dataset contains only 20 categories and is simpler than MS-COCO, the advantage of our framework over other baseline methods is less significant. However, the proposed framework demonstrates superior performance compared to other baseline methods when label proportions are extremely low (e.g., under 20\%). Specifically, it outperforms the best baseline method (i.e., DSRB) by 1.3\% in average mAP and maintains higher average OF1 and CF1 scores. Additionally, our framework consistently outperforms all other baseline methods under all label proportions.

\subsection{Analysis on Visual Semantic Decoupling}
\begin{figure*}[!t] 
	\centering 
	\includegraphics[width=1.0\linewidth]{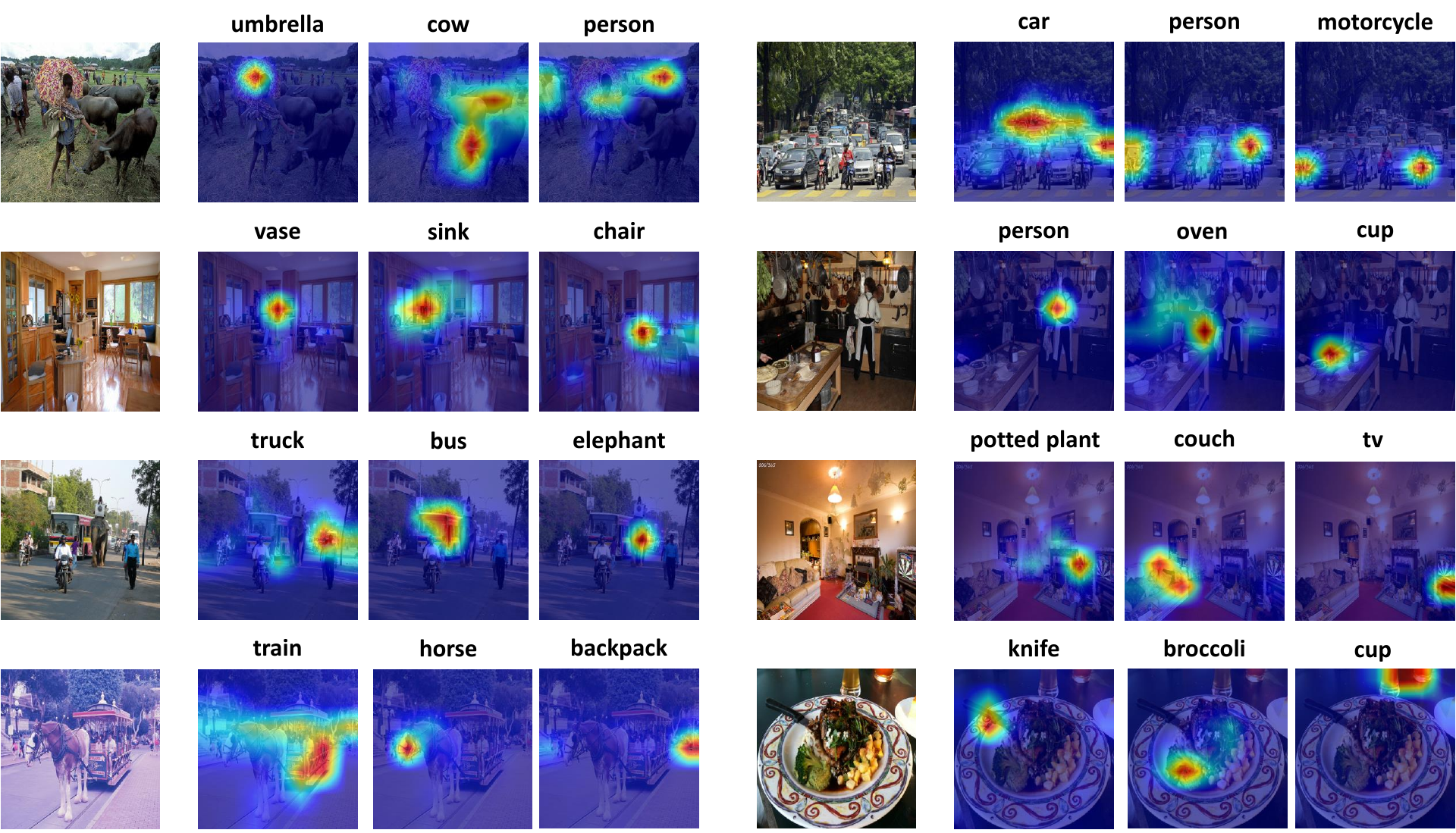}
	\caption{Several examples of input images and category activation maps corresponding to the top-3 highest confidence categories predicted by the proposed framework.}
	\label{Fig:CAM}
\end{figure*}

The semantic decoupling module is a critical component in our framework, which is introduced to extract semantic-specific information from the global representation obtained by the CLIP visual encoder. To demonstrate the effectiveness of the semantic decoupling module, we visualize examples from the MS-COCO dataset in Figure \ref{Fig:CAM}. Each row contains two images and the corresponding class activation maps (CAMs) for the top-3 highest confidence positive categories in each image. These results illustrate that the semantic decoupling module can efficiently highlight objects of the corresponding category, even when the scenes and relationships between objects in the image are complex. For example, the module precisely locates the regions of the umbrella, cow and person in the first sample, even when these objects are spatially overlapping. This capability is also evident in other samples. Thus, our framework effectively avoids semantic confusion, as the visual features corresponding to different categories of objects are distinctly separated.

\begin{figure*}[!t] 
	\centering 
	\includegraphics[width=1.0\linewidth]{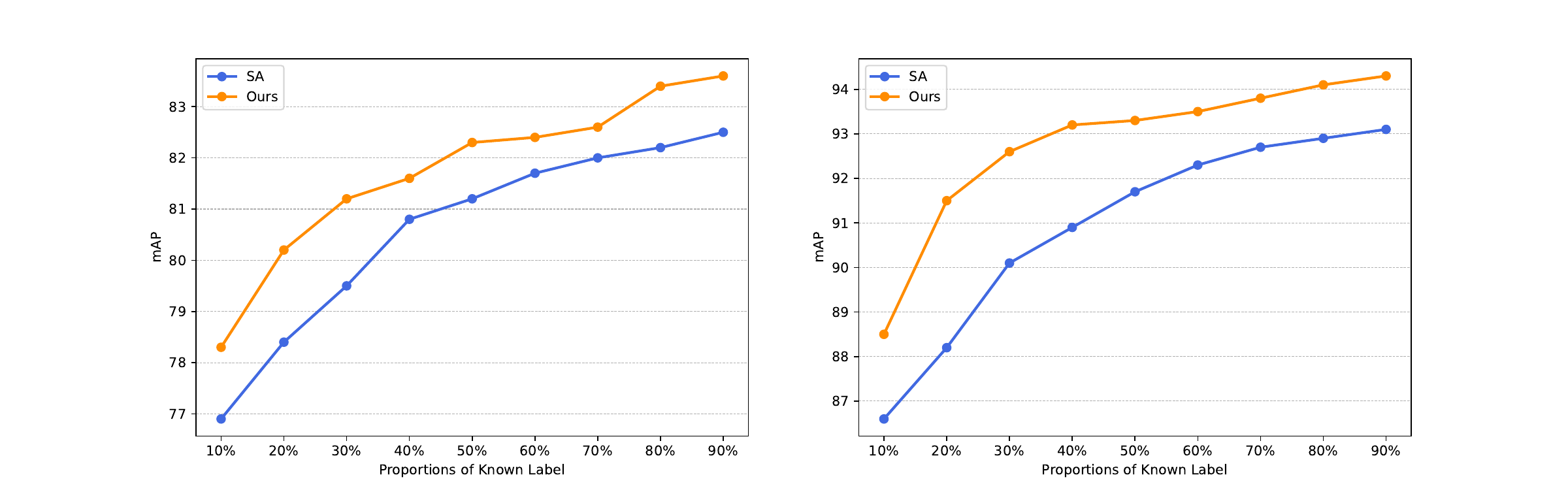}
	\caption{Comparison of the proposed semantic decoupling module and semantic attention fusion method (SA) under all label settings in the MS-COCO (left) and Pascal VOC (right) datasets.}
	\label{Fig:variant}
\end{figure*}
Additionally, there are alternative methods that can be utilized within the MLR framework to extract fine-grained visual features for different categories. To further demonstrate the superiority of semantic decoupling, we investigate additional design variants for the decoupling of global visual representation within the CLIP-based framework. Semantic attention (SA) \cite{ye2020SA} is a prominent method that utilizes stacked convolutional layers to extract content-aware feature maps for each category. For a fair comparison, we replace the proposed semantic decoupling module with the SA module while maintaining the rest of the framework unchanged and conduct experiments across all settings. As illustrated in Figure \ref{Fig:variant}, the semantic decoupling method demonstrates significant advantages, achieving average mAP improvements of 1.1\% and 1.9\% on the MS-COCO and Pascal VOC datasets, respectively. These results demonstrate the effectiveness of the proposed semantic decoupling framework, which incorporates prior knowledge from the pre-trained semantic model into the network architecture to more effectively guide the decoupling of spatial feature information across different categories, thereby improving the performance of CLIP-based methods.

\subsection{Ablation Studies}
Since our method builds on CoOp \cite{zhou2022coop}, we emphasize comparisons with CoOp to evaluate the effectiveness of our framework. We conduct experiments both on MS-COCO and Pascal VOC datasets, under the label proportions of 10\%, 30\%, 50\%, 70\% and 90\%. 

We first evaluate the contribution of the semantic decoupling module. As shown in Table \ref{Ablation}, the baseline CoOp method obtains average mAP values of 74.6\%, 89.9\% on MS-COCO dataset and Pascal VOC datasets, respectively. By adding the semantic decoupling module to the end of CLIP visual encoder (CoOp w/sd), the average mAP values are boosted to 80.1\% and 91.8\%, with improvements of 5.5\% and 1.9\%, respectively. We further introduce the category-specific prompt optimization method, considering the fine-grained property of MLR. To evaluate its effectiveness, we also conduct experiments by adding it to the baseline method (CoOp w/csp). As presented in Table \ref{Ablation}, the mAP values of CoOp improves to 76.9\% and 90.6\% when utilizing the category-specific prompt optimization method. 
Additionally, incorporating the prompt optimization method on top of the semantic decoupling module integration can further improve the mAP values by 1.5\% and 0.7\%, respectively.

\begin{table*}[!ht] 
    \centering 
    \small
    \caption{mAP comparisons between the baseline CoOp method, CoOp with category-specific prompt optimization (CoOp w/csp), CoOp with semantic decoupling module (CoOp w/sd) and our framework (Ours) incorporating both category-specific prompt optimization and semantic decoupling module on the MS-COCO and Pascal VOC 2007 datasets under certain proportions of known labels.}
    \vspace{5pt}
    \scalebox{1.0}{
    \begin{tabular*}{\textwidth}{@{\extracolsep{\fill}} c|c|ccccc|c}
    \hline
    Datasets &Methods &10\% &30\% &50\% &70\% &90\% &Avg.\\ 
    \hline \noalign{\smallskip} \hline
    \multirow{4}{*}{\centering MS-COCO}
    &CoOp \cite{zhou2022coop}  &71.2 &73.0 &75.6 &76.3 &76.9 &74.6\\
    &CoOp w/csp &72.3 &76.9 &77.8 &78.4 &78.9 &76.9\\
    &CoOp w/sd &76.7 &79.8 &80.5 &81.2 &82.1 &80.1 \\
    &\textbf{Ours} &\textbf{78.3} &\textbf{81.2} &\textbf{82.3} &\textbf{82.6} &\textbf{83.6} &\textbf{81.6}\\
    \hline \noalign{\smallskip} \hline
    \multirow{4}{*}{\centering Pascal VOC}
    &CoOp \cite{zhou2022coop}  &85.7 &88.9 &91.3 &91.9 &92.0 &89.9\\
    &CoOp w/csp &86.3 &89.3 &91.8 &92.6 &93.2 &90.6 \\
    &CoOp w/sd &87.2 &91.7 &92.9 &93.4 &94.0 &91.8\\
    &\textbf{Ours} &\textbf{88.5} &\textbf{92.6} &\textbf{93.3} &\textbf{93.8} &\textbf{94.3} &\textbf{92.5}\\
    \hline
\end{tabular*}
}
\label{Ablation} 
\end{table*}

\begin{figure*}[!htbp] 
	\centering 
	\includegraphics[width=1.0\linewidth]{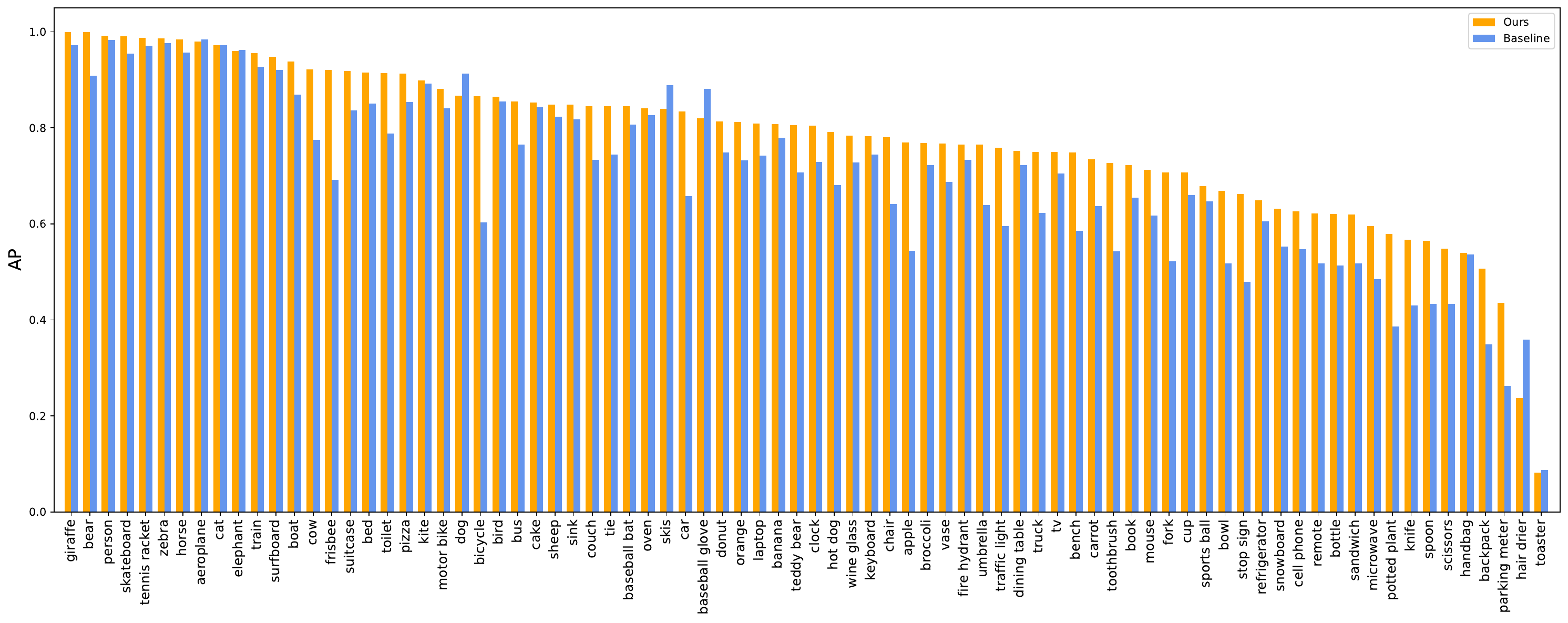}
	\caption{Per-class average precision (AP) of our proposed framework and baseline method with known label proportion of 10\% on the MS-COCO dataset.}
	\label{Fig:Ap-perclass}
\end{figure*}

To further analyze the improvements compared to the baseline method, we present the average precision (AP) of each category when the known label proportion is 10\% in the MS-COCO dataset, as shown in Figure \ref{Fig:Ap-perclass}. It demonstrates that the our method significantly improves classification performance in most categories compared to the baseline method.
\begin{figure*}[!t] 
	\centering 
	\includegraphics[width=1.0\linewidth]{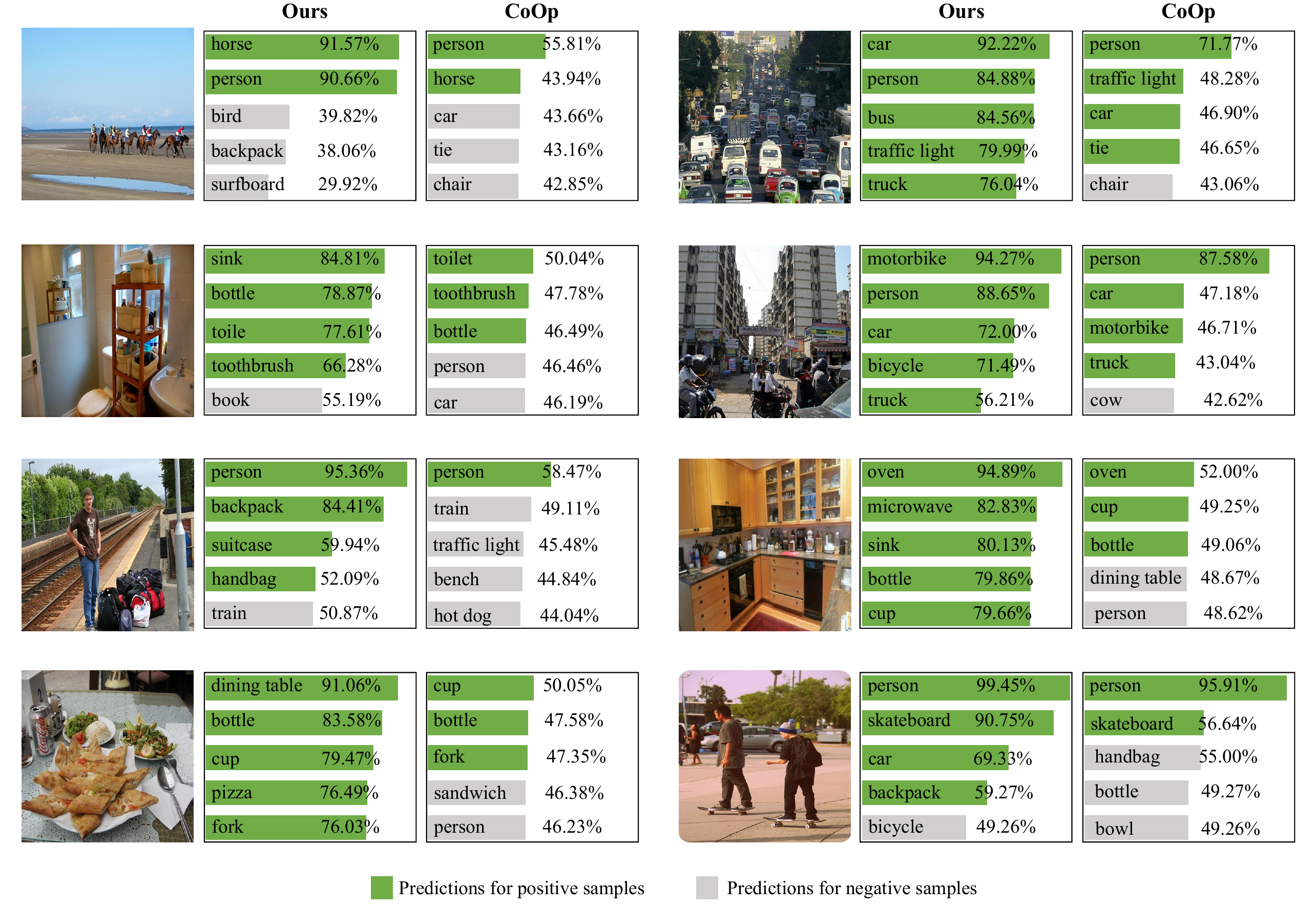}
	\caption{Visualizations of multi-label classification results for several images predicted by our proposed method and the baseline CoOp method. Both models are trained on the MS-COCO dataset with known label proportion of 10\%.}
	\label{Fig:illustration}
\end{figure*}
Moreover, we present some classification examples from the MS-COCO dataset under the same setting, as shown in Figure \ref{Fig:illustration}. We compare the top-5 highest scores of our method with those of the baseline method in each image. These results illustrate that our framework achieves comprehensive recognition results with high accuracy even in complex scenes. In contrast, the baseline method exhibits less distinguishable scores for different categories in complex scenarios, which degrades its performance.
For example, in the first sample of the third row, our method successfully recognizes all positive labels, whereas the baseline method only recognizes the 'person' category. Similar situations are observed in other examples, where our method consistently achieves higher prediction scores for the positive categories compared to the baseline method. Despite the small size of objects and their complex interactions within the scene, our method maintains excellent performance due to the effectiveness of visual semantic decoupling mechanism, which separates the object information with different semantics.

\begin{figure*}[!htbp] 
	\centering 
	\includegraphics[width=1.0\linewidth]{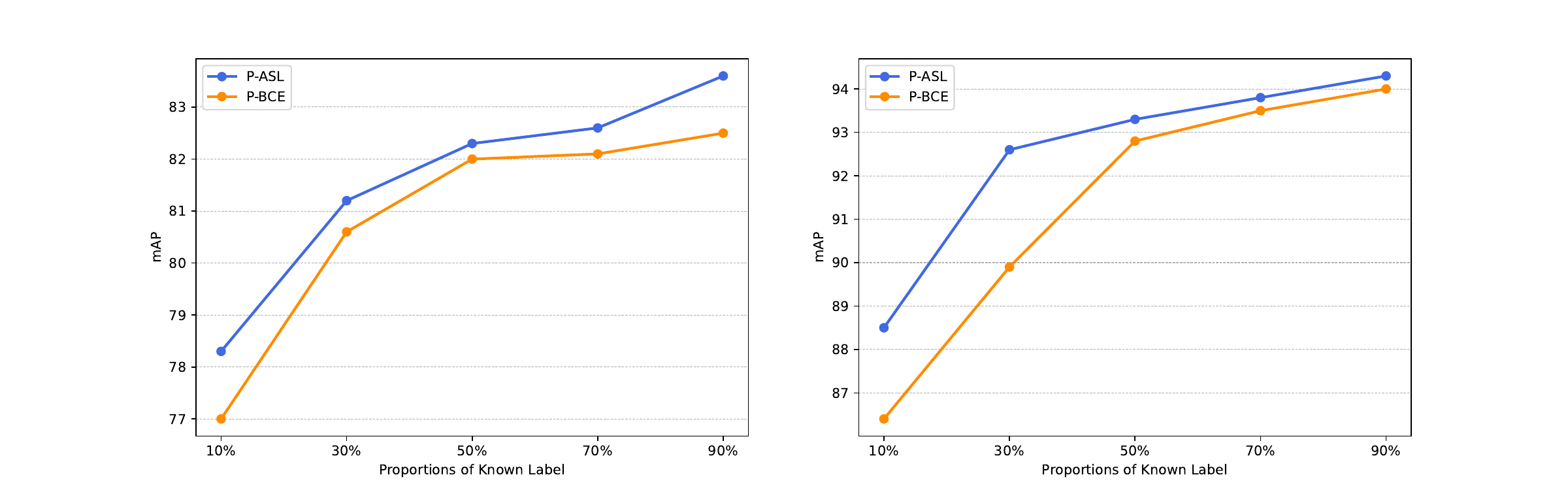}
	\caption{Comparison of P-ASL and P-BCE as loss functions for the proposed framework. All the experiments are conducted on the MS-COCO dataset (left) and Pascal VOC dataset (right) with known label proportions from 10\% to 90\%.}
	\label{Fig:ASLandBCE}
\end{figure*}
Except the above discussion, we further conduct experiments to explore the benefits of P-ASL on our task, which is adopted to alleviate the problem of the imbalance between negative and positive samples in MLR-PL. Such imbalance is overlooked in traditional partial BCE loss (P-BCE) \cite{durand2019learning}. Specifically, we alternated between P-ASL and P-BCE as the loss function while keeping the rest of the framework unchanged. As shown in Figure \ref{Fig:ASLandBCE}, using P-ASL results in an average mAP improvement of 1.1\% on MS-COCO and 1.2\% on Pascal VOC compared to P-BCE, respectively. Moreover, P-ASL demonstrates significant advantages when label proportions are particularly low (below 30\%), where the negative-positive imbalance becomes more severe. These results suggest that adopting P-ASL effectively mitigates the optimization problems caused by this imbalance in partial label settings.

\begin{figure*}[!htbp] 
	\centering 
	\includegraphics[width=1.0\linewidth]{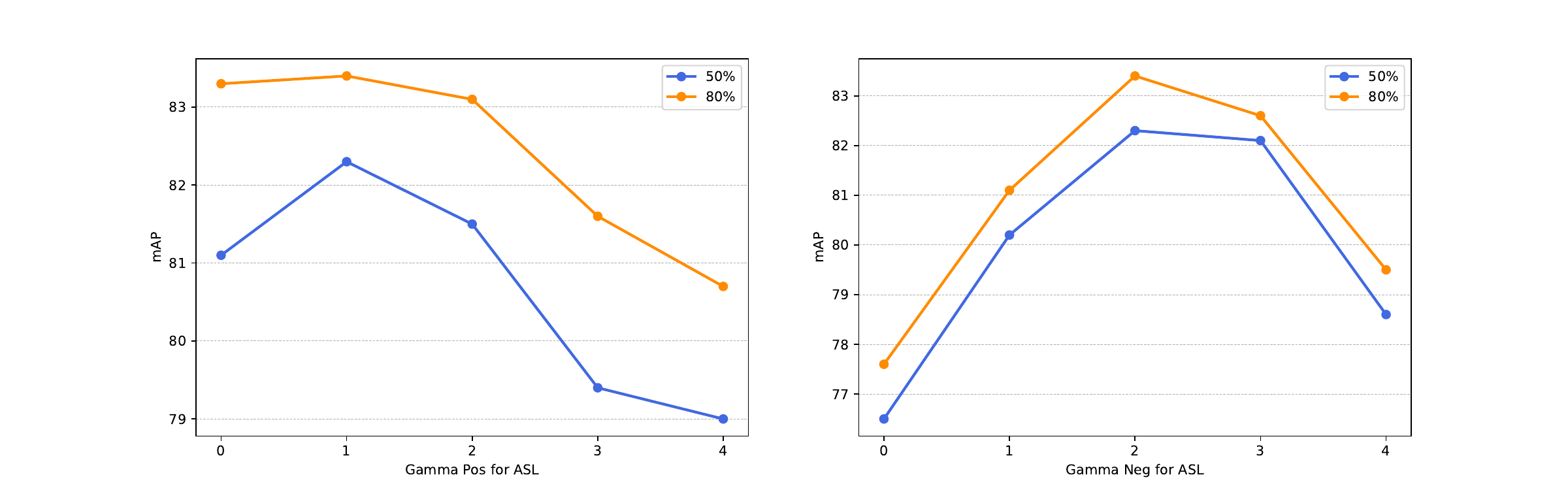}
	\caption{Analysis of the effect of $\gamma_{+}$ (left) and $\gamma_{-}$ (right) in P-ASL. These experiments are conducted on the MS-COCO dataset with known label proportions of 50\% and 80\%.}
	\label{Fig:gamma}
\end{figure*}
Moreover, the parameters $\gamma_{+}$ and $\gamma_{-}$ are critical to the performance of P-ASL. To gain a more comprehensive understanding, we further explore the optimal parameter choices in our framework. In Figure \ref{Fig:gamma}, we present the variation in mAP as $\gamma_{+}$ changes from 0 to 4 under the 50\% and 80\% known label settings, with $\gamma_{-}$ fixed at 2. The framework achieves its best performance at $\gamma_{+}$ = 1 in both settings, while further increases in $\gamma_{+}$ lead to a drop in performance. We also analyze the effect of $\gamma_{-}$ while keeping $\gamma_{+}$ fixed at 1. As $\gamma_{-}$ increases from 0 to 2, the mAP improves from 76.5 and 77.6 to 82.0 and 82.3, respectively, but drops sharply with further increases in $\gamma_{-}$. Thus, we set $\gamma_{+}$ = 1 and $\gamma_{-}$ = 2 as the optimal parameters for P-ASL in our framework.

\begin{figure*}[!tbp] 
	\centering 
	\includegraphics[width=1.0\linewidth]{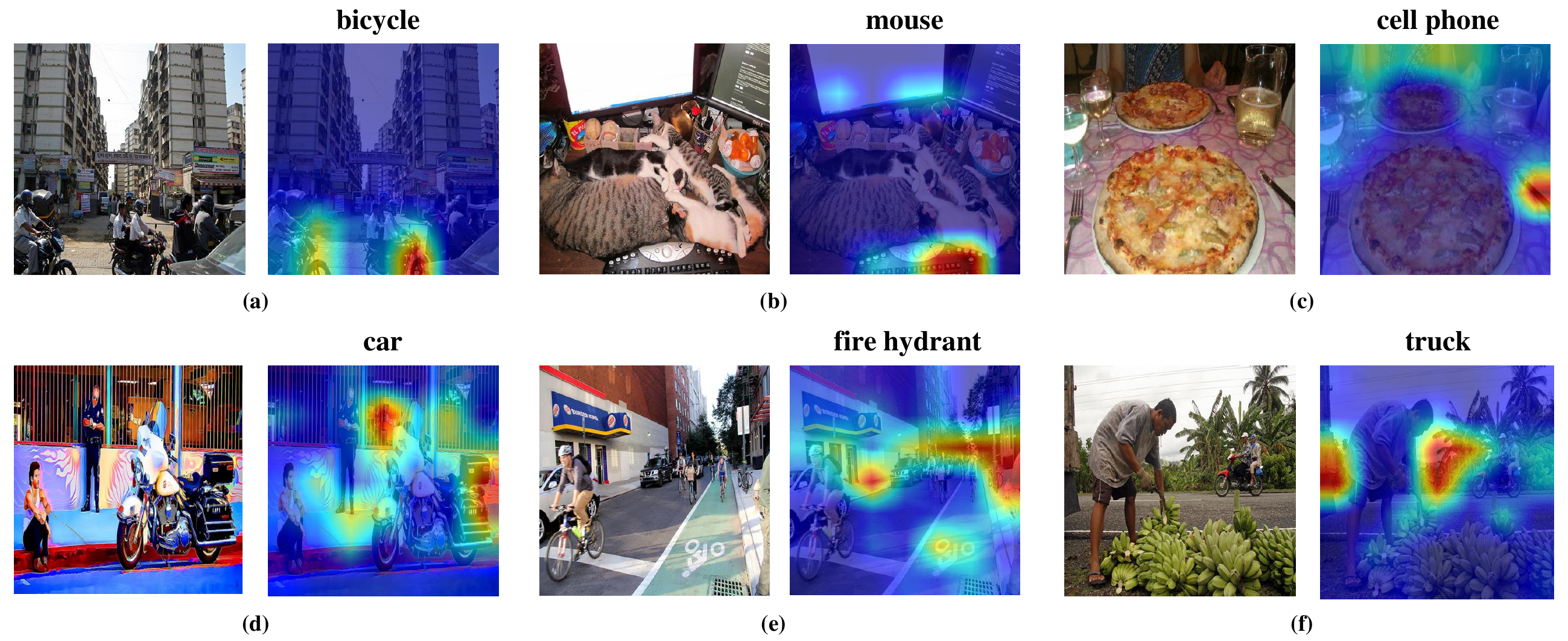}
	\caption{Examples of failure cases in the proposed semantic decoupling method. Typical failures include inaccurate positioning (a, b, c) and excessive positioning (d, e, f) of specific objects in the attention map.}
	\label{Fig:Failure}
\end{figure*}

\subsection{Limitations}

Despite the effectiveness of the semantic decoupling method in separating information from different semantic objects, it still exhibits certain limitations, particularly when objects in an image are small or severely occluded. Several typical failure cases are illustrated in Figure \ref{Fig:Failure}. The localization of the attention map may lack accuracy when objects are occluded. For instance, the motorcycle is incorrectly recognized as a bicycle due to semantic similarity in (a), as the bicycle is largely overlapped. Similarly, the mouse and cell phone in (b) and (c) are misclassified as other relevant semantic targets due to their lack of prominence. Additional failure cases in the attention map occur when the area of the objects is excessively large and contains background noise. For example, the attention map for the car in (d) extends beyond the relevant area and covers part of the motorcycle. Similar issues are observed in (e) and (f), where the fire hydrant and truck are less salient, leading to uncertainty in the positioning of the target areas. These problems occur when information about small targets is overlooked in the global representation after coarse-grained operations, resulting in the failure to effectively align semantic features with the corresponding spatial feature regions. Nonetheless, in images containing multiple targets, these decoupled features remain valuable, as they incorporate less background noise and mixed target information compared to the global representation.

\section{Conclusions}

In this work, we introduce a CLIP-based framework to address the multi-label recognition with partial labels (MLR-PL) task, in which only some labels are known while others are unknown in each image. 
We propose a semantic-guided spatial attention mechanism and a category-specific prompt optimization method to learn decoupled visual and textual features for each category. By integrating these components, our framework captures fine-grained information for predicting different categories, thereby improving the classification accuracy in MLR-PL. 
Extensive experiments on the MS-COCO 2014 and Pascal VOC 2007 datasets demonstrate the superiority of our framework compared to the current leading algorithms. Additionally, visual analysis shows that our method can effectively decouple feature information of different categories.
In the future, we will continue to enhance the semantic decoupling framework by introducing label correlation information, which can improve the performance on MLR-PL task by enriching the contextual semantic information in categorical features.
\begin{acks}
This work was supported in part by National Natural Science Foundation of China (NSFC) under Grant No. 62206060 and No. 62206314, in Part by Natural Science Foundation of Guangdong Province (2022A1515011555, 2023A1515012568, 2023A1515012561), Guangdong Provincial Key Laboratory of Human Digital Twin (2022B1212010004), in part by Science and Technology Projects in Guangzhou under Grant No. 2024A04J4388, and in part by Guangzhou Basic and Applied Basic Research Foundation under Grant No. SL2022A04J01626.
\end{acks}

\bibliographystyle{ACM-Reference-Format}
\bibliography{sample-base}


\begin{thebibliography}{56}


\ifx \showCODEN    \undefined \def \showCODEN     #1{\unskip}     \fi
\ifx \showDOI      \undefined \def \showDOI       #1{#1}\fi
\ifx \showISBNx    \undefined \def \showISBNx     #1{\unskip}     \fi
\ifx \showISBNxiii \undefined \def \showISBNxiii  #1{\unskip}     \fi
\ifx \showISSN     \undefined \def \showISSN      #1{\unskip}     \fi
\ifx \showLCCN     \undefined \def \showLCCN      #1{\unskip}     \fi
\ifx \shownote     \undefined \def \shownote      #1{#1}          \fi
\ifx \showarticletitle \undefined \def \showarticletitle #1{#1}   \fi
\ifx \showURL      \undefined \def \showURL       {\relax}        \fi
\providecommand\bibfield[2]{#2}
\providecommand\bibinfo[2]{#2}
\providecommand\natexlab[1]{#1}
\providecommand\showeprint[2][]{arXiv:#2}

\bibitem[Bulat and Tzimiropoulos(2023)]%
        {bulat2023lasprompt}
\bibfield{author}{\bibinfo{person}{Adrian Bulat} {and} \bibinfo{person}{Georgios Tzimiropoulos}.} \bibinfo{year}{2023}\natexlab{}.
\newblock \showarticletitle{Lasp: Text-to-text optimization for language-aware soft prompting of vision \& language models}. In \bibinfo{booktitle}{\emph{Proceedings of the IEEE/CVF Conference on Computer Vision and Pattern Recognition}}. \bibinfo{pages}{23232--23241}.
\newblock


\bibitem[Carrillo et~al\mbox{.}(2013)]%
        {carrillo2013multi}
\bibfield{author}{\bibinfo{person}{Dolly Carrillo}, \bibinfo{person}{Vivian~F L{\'o}pez}, {and} \bibinfo{person}{Mar{\'\i}a~N Moreno}.} \bibinfo{year}{2013}\natexlab{}.
\newblock \showarticletitle{Multi-label classification for recommender systems}. In \bibinfo{booktitle}{\emph{Trends in Practical Applications of Agents and Multiagent Systems: 11th International Conference on Practical Applications of Agents and Multi-Agent Systems}}. Springer, \bibinfo{pages}{181--188}.
\newblock


\bibitem[Chen et~al\mbox{.}(2022a)]%
        {chen2022knowledge}
\bibfield{author}{\bibinfo{person}{Tianshui Chen}, \bibinfo{person}{Liang Lin}, \bibinfo{person}{Riquan Chen}, \bibinfo{person}{Xiaolu Hui}, {and} \bibinfo{person}{Hefeng Wu}.} \bibinfo{year}{2022}\natexlab{a}.
\newblock \showarticletitle{Knowledge-Guided Multi-Label Few-Shot Learning for General Image Recognition}.
\newblock \bibinfo{journal}{\emph{IEEE Transactions on Pattern Analysis and Machine Intelligence}} \bibinfo{volume}{44}, \bibinfo{number}{3} (\bibinfo{year}{2022}), \bibinfo{pages}{1371--1384}.
\newblock
\urldef\tempurl%
\url{https://doi.org/10.1109/TPAMI.2020.3025814}
\showDOI{\tempurl}


\bibitem[Chen et~al\mbox{.}(2024a)]%
        {Chen2024HST}
\bibfield{author}{\bibinfo{person}{Tianshui Chen}, \bibinfo{person}{Tao Pu}, \bibinfo{person}{Lingbo Liu}, \bibinfo{person}{Yukai Shi}, \bibinfo{person}{Zhijing Yang}, {and} \bibinfo{person}{Liang Lin}.} \bibinfo{year}{2024}\natexlab{a}.
\newblock \showarticletitle{Heterogeneous semantic transfer for multi-label recognition with partial labels}.
\newblock \bibinfo{journal}{\emph{International Journal of Computer Vision}} (\bibinfo{year}{2024}), \bibinfo{pages}{1--16}.
\newblock


\bibitem[Chen et~al\mbox{.}(2022b)]%
        {chen2022structured}
\bibfield{author}{\bibinfo{person}{Tianshui Chen}, \bibinfo{person}{Tao Pu}, \bibinfo{person}{Hefeng Wu}, \bibinfo{person}{Yuan Xie}, {and} \bibinfo{person}{Liang Lin}.} \bibinfo{year}{2022}\natexlab{b}.
\newblock \showarticletitle{Structured semantic transfer for multi-label recognition with partial labels}. In \bibinfo{booktitle}{\emph{Proceedings of the AAAI conference on artificial intelligence}}, Vol.~\bibinfo{volume}{36}. \bibinfo{pages}{339--346}.
\newblock


\bibitem[Chen et~al\mbox{.}(2021)]%
        {chen2021cross}
\bibfield{author}{\bibinfo{person}{Tianshui Chen}, \bibinfo{person}{Tao Pu}, \bibinfo{person}{Hefeng Wu}, \bibinfo{person}{Yuan Xie}, \bibinfo{person}{Lingbo Liu}, {and} \bibinfo{person}{Liang Lin}.} \bibinfo{year}{2021}\natexlab{}.
\newblock \showarticletitle{Cross-domain facial expression recognition: A unified evaluation benchmark and adversarial graph learning}.
\newblock \bibinfo{journal}{\emph{IEEE transactions on pattern analysis and machine intelligence}} \bibinfo{volume}{44}, \bibinfo{number}{12} (\bibinfo{year}{2021}), \bibinfo{pages}{9887--9903}.
\newblock


\bibitem[Chen et~al\mbox{.}(2024b)]%
        {chen2024dynamic}
\bibfield{author}{\bibinfo{person}{Tianshui Chen}, \bibinfo{person}{Weihang Wang}, \bibinfo{person}{Tao Pu}, \bibinfo{person}{Jinghui Qin}, \bibinfo{person}{Zhijing Yang}, \bibinfo{person}{Jie Liu}, {and} \bibinfo{person}{Liang Lin}.} \bibinfo{year}{2024}\natexlab{b}.
\newblock \showarticletitle{Dynamic Correlation Learning and Regularization for Multi-Label Confidence Calibration}.
\newblock \bibinfo{journal}{\emph{arXiv preprint arXiv:2407.06844}} (\bibinfo{year}{2024}).
\newblock


\bibitem[Chen et~al\mbox{.}(2018)]%
        {chen2018recurrent}
\bibfield{author}{\bibinfo{person}{Tianshui Chen}, \bibinfo{person}{Zhouxia Wang}, \bibinfo{person}{Guanbin Li}, {and} \bibinfo{person}{Liang Lin}.} \bibinfo{year}{2018}\natexlab{}.
\newblock \showarticletitle{Recurrent Attentional Reinforcement Learning for Multi-Label Image Recognition}.
\newblock \bibinfo{journal}{\emph{Proceedings of the AAAI Conference on Artificial Intelligence}} \bibinfo{volume}{32}, \bibinfo{number}{1} (\bibinfo{date}{Apr.} \bibinfo{year}{2018}).
\newblock
\urldef\tempurl%
\url{https://doi.org/10.1609/aaai.v32i1.12281}
\showDOI{\tempurl}


\bibitem[Chen et~al\mbox{.}(2019b)]%
        {chen2019learning}
\bibfield{author}{\bibinfo{person}{Tianshui Chen}, \bibinfo{person}{Muxin Xu}, \bibinfo{person}{Xiaolu Hui}, \bibinfo{person}{Hefeng Wu}, {and} \bibinfo{person}{Liang Lin}.} \bibinfo{year}{2019}\natexlab{b}.
\newblock \showarticletitle{Learning Semantic-Specific Graph Representation for Multi-Label Image Recognition}. In \bibinfo{booktitle}{\emph{Proceedings of the IEEE/CVF International Conference on Computer Vision (ICCV)}}.
\newblock


\bibitem[Chen et~al\mbox{.}(2019a)]%
        {chen2019multi}
\bibfield{author}{\bibinfo{person}{Zhao-Min Chen}, \bibinfo{person}{Xiu-Shen Wei}, \bibinfo{person}{Peng Wang}, {and} \bibinfo{person}{Yanwen Guo}.} \bibinfo{year}{2019}\natexlab{a}.
\newblock \showarticletitle{Multi-label image recognition with graph convolutional networks}. In \bibinfo{booktitle}{\emph{Proceedings of the IEEE/CVF conference on computer vision and pattern recognition}}. \bibinfo{pages}{5177--5186}.
\newblock


\bibitem[Cheng et~al\mbox{.}(2005)]%
        {cheng2005semantic}
\bibfield{author}{\bibinfo{person}{Shyi-Chyi Cheng}, \bibinfo{person}{Tzu-Chuan Chou}, \bibinfo{person}{Chao-Lung Yang}, {and} \bibinfo{person}{Hung-Yi Chang}.} \bibinfo{year}{2005}\natexlab{}.
\newblock \showarticletitle{A semantic learning for content-based image retrieval using analytical hierarchy process}.
\newblock \bibinfo{journal}{\emph{Expert Systems with Applications}} \bibinfo{volume}{28}, \bibinfo{number}{3} (\bibinfo{year}{2005}), \bibinfo{pages}{495--505}.
\newblock


\bibitem[Darban and Valipour(2022)]%
        {darban2022ghrs}
\bibfield{author}{\bibinfo{person}{Zahra~Zamanzadeh Darban} {and} \bibinfo{person}{Mohammad~Hadi Valipour}.} \bibinfo{year}{2022}\natexlab{}.
\newblock \showarticletitle{GHRS: Graph-based hybrid recommendation system with application to movie recommendation}.
\newblock \bibinfo{journal}{\emph{Expert Systems with Applications}}  \bibinfo{volume}{200} (\bibinfo{year}{2022}), \bibinfo{pages}{116850}.
\newblock


\bibitem[Dosovitskiy et~al\mbox{.}(2021)]%
        {dosovitskiy2020vit}
\bibfield{author}{\bibinfo{person}{Alexey Dosovitskiy}, \bibinfo{person}{Lucas Beyer}, \bibinfo{person}{Alexander Kolesnikov}, \bibinfo{person}{Dirk Weissenborn}, \bibinfo{person}{Xiaohua Zhai}, \bibinfo{person}{Thomas Unterthiner}, \bibinfo{person}{Mostafa Dehghani}, \bibinfo{person}{Matthias Minderer}, \bibinfo{person}{Georg Heigold}, \bibinfo{person}{Sylvain Gelly}, \bibinfo{person}{Jakob Uszkoreit}, {and} \bibinfo{person}{Neil Houlsby}.} \bibinfo{year}{2021}\natexlab{}.
\newblock \showarticletitle{An Image is Worth 16x16 Words: Transformers for Image Recognition at Scale}. In \bibinfo{booktitle}{\emph{International Conference on Learning Representations}}.
\newblock
\urldef\tempurl%
\url{https://openreview.net/forum?id=YicbFdNTTy}
\showURL{%
\tempurl}


\bibitem[Du et~al\mbox{.}(2022)]%
        {du2022openvoca}
\bibfield{author}{\bibinfo{person}{Yu Du}, \bibinfo{person}{Fangyun Wei}, \bibinfo{person}{Zihe Zhang}, \bibinfo{person}{Miaojing Shi}, \bibinfo{person}{Yue Gao}, {and} \bibinfo{person}{Guoqi Li}.} \bibinfo{year}{2022}\natexlab{}.
\newblock \showarticletitle{Learning to prompt for open-vocabulary object detection with vision-language model}. In \bibinfo{booktitle}{\emph{Proceedings of the IEEE/CVF Conference on Computer Vision and Pattern Recognition}}. \bibinfo{pages}{14084--14093}.
\newblock


\bibitem[Durand et~al\mbox{.}(2019)]%
        {durand2019learning}
\bibfield{author}{\bibinfo{person}{Thibaut Durand}, \bibinfo{person}{Nazanin Mehrasa}, {and} \bibinfo{person}{Greg Mori}.} \bibinfo{year}{2019}\natexlab{}.
\newblock \showarticletitle{Learning a deep convnet for multi-label classification with partial labels}. In \bibinfo{booktitle}{\emph{Proceedings of the IEEE/CVF conference on computer vision and pattern recognition}}. \bibinfo{pages}{647--657}.
\newblock


\bibitem[Everingham et~al\mbox{.}(2010)]%
        {everingham2010voc}
\bibfield{author}{\bibinfo{person}{Mark Everingham}, \bibinfo{person}{Luc Van~Gool}, \bibinfo{person}{Christopher~KI Williams}, \bibinfo{person}{John Winn}, {and} \bibinfo{person}{Andrew Zisserman}.} \bibinfo{year}{2010}\natexlab{}.
\newblock \showarticletitle{The pascal visual object classes (voc) challenge}.
\newblock \bibinfo{journal}{\emph{International journal of computer vision}}  \bibinfo{volume}{88} (\bibinfo{year}{2010}), \bibinfo{pages}{303--338}.
\newblock


\bibitem[Ge et~al\mbox{.}(2023)]%
        {DAprompt}
\bibfield{author}{\bibinfo{person}{Chunjiang Ge}, \bibinfo{person}{Rui Huang}, \bibinfo{person}{Mixue Xie}, \bibinfo{person}{Zihang Lai}, \bibinfo{person}{Shiji Song}, \bibinfo{person}{Shuang Li}, {and} \bibinfo{person}{Gao Huang}.} \bibinfo{year}{2023}\natexlab{}.
\newblock \showarticletitle{Domain Adaptation via Prompt Learning}.
\newblock \bibinfo{journal}{\emph{IEEE Transactions on Neural Networks and Learning Systems}} (\bibinfo{year}{2023}), \bibinfo{pages}{1--11}.
\newblock
\urldef\tempurl%
\url{https://doi.org/10.1109/TNNLS.2023.3327962}
\showDOI{\tempurl}


\bibitem[He et~al\mbox{.}(2016)]%
        {he2016resnet}
\bibfield{author}{\bibinfo{person}{Kaiming He}, \bibinfo{person}{Xiangyu Zhang}, \bibinfo{person}{Shaoqing Ren}, {and} \bibinfo{person}{Jian Sun}.} \bibinfo{year}{2016}\natexlab{}.
\newblock \showarticletitle{Deep residual learning for image recognition}. In \bibinfo{booktitle}{\emph{Proceedings of the IEEE conference on computer vision and pattern recognition}}. \bibinfo{pages}{770--778}.
\newblock


\bibitem[He et~al\mbox{.}(2023)]%
        {He2023openvoca}
\bibfield{author}{\bibinfo{person}{Sunan He}, \bibinfo{person}{Taian Guo}, \bibinfo{person}{Tao Dai}, \bibinfo{person}{Ruizhi Qiao}, \bibinfo{person}{Xiujun Shu}, \bibinfo{person}{Bo Ren}, {and} \bibinfo{person}{Shu-Tao Xia}.} \bibinfo{year}{2023}\natexlab{}.
\newblock \showarticletitle{Open-Vocabulary Multi-Label Classification via Multi-Modal Knowledge Transfer}.
\newblock \bibinfo{journal}{\emph{Proceedings of the AAAI Conference on Artificial Intelligence}} \bibinfo{volume}{37}, \bibinfo{number}{1} (\bibinfo{date}{Jun.} \bibinfo{year}{2023}), \bibinfo{pages}{808--816}.
\newblock
\urldef\tempurl%
\url{https://doi.org/10.1609/aaai.v37i1.25159}
\showDOI{\tempurl}


\bibitem[Hochreiter and Schmidhuber(1997)]%
        {hochreiter1997long}
\bibfield{author}{\bibinfo{person}{Sepp Hochreiter} {and} \bibinfo{person}{J{\"u}rgen Schmidhuber}.} \bibinfo{year}{1997}\natexlab{}.
\newblock \showarticletitle{Long short-term memory}.
\newblock \bibinfo{journal}{\emph{Neural computation}} \bibinfo{volume}{9}, \bibinfo{number}{8} (\bibinfo{year}{1997}), \bibinfo{pages}{1735--1780}.
\newblock


\bibitem[Huynh and Elhamifar(2020)]%
        {huynh2020interactive}
\bibfield{author}{\bibinfo{person}{Dat Huynh} {and} \bibinfo{person}{Ehsan Elhamifar}.} \bibinfo{year}{2020}\natexlab{}.
\newblock \showarticletitle{Interactive multi-label cnn learning with partial labels}. In \bibinfo{booktitle}{\emph{Proceedings of the IEEE/CVF Conference on Computer Vision and Pattern Recognition}}. \bibinfo{pages}{9423--9432}.
\newblock


\bibitem[Jia et~al\mbox{.}(2021)]%
        {jia2021scaling}
\bibfield{author}{\bibinfo{person}{Chao Jia}, \bibinfo{person}{Yinfei Yang}, \bibinfo{person}{Ye Xia}, \bibinfo{person}{Yi-Ting Chen}, \bibinfo{person}{Zarana Parekh}, \bibinfo{person}{Hieu Pham}, \bibinfo{person}{Quoc Le}, \bibinfo{person}{Yun-Hsuan Sung}, \bibinfo{person}{Zhen Li}, {and} \bibinfo{person}{Tom Duerig}.} \bibinfo{year}{2021}\natexlab{}.
\newblock \showarticletitle{Scaling Up Visual and Vision-Language Representation Learning With Noisy Text Supervision}. In \bibinfo{booktitle}{\emph{Proceedings of the 38th International Conference on Machine Learning}} \emph{(\bibinfo{series}{Proceedings of Machine Learning Research}, Vol.~\bibinfo{volume}{139})}, \bibfield{editor}{\bibinfo{person}{Marina Meila} {and} \bibinfo{person}{Tong Zhang}} (Eds.). \bibinfo{publisher}{PMLR}, \bibinfo{pages}{4904--4916}.
\newblock
\urldef\tempurl%
\url{https://proceedings.mlr.press/v139/jia21b.html}
\showURL{%
\tempurl}


\bibitem[Jia et~al\mbox{.}(2022)]%
        {vlmprompt}
\bibfield{author}{\bibinfo{person}{Menglin Jia}, \bibinfo{person}{Luming Tang}, \bibinfo{person}{Bor-Chun Chen}, \bibinfo{person}{Claire Cardie}, \bibinfo{person}{Serge Belongie}, \bibinfo{person}{Bharath Hariharan}, {and} \bibinfo{person}{Ser-Nam Lim}.} \bibinfo{year}{2022}\natexlab{}.
\newblock \showarticletitle{Visual Prompt Tuning}. In \bibinfo{booktitle}{\emph{Computer Vision -- ECCV 2022}}, \bibfield{editor}{\bibinfo{person}{Shai Avidan}, \bibinfo{person}{Gabriel Brostow}, \bibinfo{person}{Moustapha Ciss{\'e}}, \bibinfo{person}{Giovanni~Maria Farinella}, {and} \bibinfo{person}{Tal Hassner}} (Eds.). \bibinfo{publisher}{Springer Nature Switzerland}, \bibinfo{address}{Cham}, \bibinfo{pages}{709--727}.
\newblock
\showISBNx{978-3-031-19827-4}


\bibitem[Kim et~al\mbox{.}(2023a)]%
        {kim2023regionopenvoca}
\bibfield{author}{\bibinfo{person}{Dahun Kim}, \bibinfo{person}{Anelia Angelova}, {and} \bibinfo{person}{Weicheng Kuo}.} \bibinfo{year}{2023}\natexlab{a}.
\newblock \showarticletitle{Region-aware pretraining for open-vocabulary object detection with vision transformers}. In \bibinfo{booktitle}{\emph{Proceedings of the IEEE/CVF conference on computer vision and pattern recognition}}. \bibinfo{pages}{11144--11154}.
\newblock


\bibitem[Kim et~al\mbox{.}(2017)]%
        {kim2016hadamard}
\bibfield{author}{\bibinfo{person}{Jin-Hwa Kim}, \bibinfo{person}{Kyoung-Woon On}, \bibinfo{person}{Woosang Lim}, \bibinfo{person}{Jeonghee Kim}, \bibinfo{person}{Jung-Woo Ha}, {and} \bibinfo{person}{Byoung-Tak Zhang}.} \bibinfo{year}{2017}\natexlab{}.
\newblock \showarticletitle{Hadamard Product for Low-rank Bilinear Pooling}. In \bibinfo{booktitle}{\emph{International Conference on Learning Representations}}.
\newblock
\urldef\tempurl%
\url{https://openreview.net/forum?id=r1rhWnZkg}
\showURL{%
\tempurl}


\bibitem[Kim et~al\mbox{.}(2022)]%
        {kim2022large}
\bibfield{author}{\bibinfo{person}{Youngwook Kim}, \bibinfo{person}{Jae~Myung Kim}, \bibinfo{person}{Zeynep Akata}, {and} \bibinfo{person}{Jungwoo Lee}.} \bibinfo{year}{2022}\natexlab{}.
\newblock \showarticletitle{Large loss matters in weakly supervised multi-label classification}. In \bibinfo{booktitle}{\emph{Proceedings of the IEEE/CVF Conference on Computer Vision and Pattern Recognition}}. \bibinfo{pages}{14156--14165}.
\newblock


\bibitem[Kim et~al\mbox{.}(2023b)]%
        {kim2023bridging}
\bibfield{author}{\bibinfo{person}{Youngwook Kim}, \bibinfo{person}{Jae~Myung Kim}, \bibinfo{person}{Jieun Jeong}, \bibinfo{person}{Cordelia Schmid}, \bibinfo{person}{Zeynep Akata}, {and} \bibinfo{person}{Jungwoo Lee}.} \bibinfo{year}{2023}\natexlab{b}.
\newblock \showarticletitle{Bridging the gap between model explanations in partially annotated multi-label classification}. In \bibinfo{booktitle}{\emph{Proceedings of the IEEE/CVF Conference on Computer Vision and Pattern Recognition}}. \bibinfo{pages}{3408--3417}.
\newblock


\bibitem[Kipf and Welling(2016)]%
        {kipf2016semi}
\bibfield{author}{\bibinfo{person}{Thomas~N. Kipf} {and} \bibinfo{person}{Max Welling}.} \bibinfo{year}{2016}\natexlab{}.
\newblock \showarticletitle{Semi-Supervised Classification with Graph Convolutional Networks}.
\newblock \bibinfo{journal}{\emph{CoRR}}  \bibinfo{volume}{abs/1609.02907} (\bibinfo{year}{2016}).
\newblock
\showeprint[arXiv]{1609.02907}
\urldef\tempurl%
\url{http://arxiv.org/abs/1609.02907}
\showURL{%
\tempurl}


\bibitem[Lai et~al\mbox{.}(2016)]%
        {lai2016instance}
\bibfield{author}{\bibinfo{person}{Hanjiang Lai}, \bibinfo{person}{Pan Yan}, \bibinfo{person}{Xiangbo Shu}, \bibinfo{person}{Yunchao Wei}, {and} \bibinfo{person}{Shuicheng Yan}.} \bibinfo{year}{2016}\natexlab{}.
\newblock \showarticletitle{Instance-aware hashing for multi-label image retrieval}.
\newblock \bibinfo{journal}{\emph{IEEE Transactions on Image Processing}} \bibinfo{volume}{25}, \bibinfo{number}{6} (\bibinfo{year}{2016}), \bibinfo{pages}{2469--2479}.
\newblock


\bibitem[Li et~al\mbox{.}(2010)]%
        {li2010technique}
\bibfield{author}{\bibinfo{person}{Ran Li}, \bibinfo{person}{YaFei Zhang}, \bibinfo{person}{Zining Lu}, \bibinfo{person}{Jianjiang Lu}, {and} \bibinfo{person}{Yulong Tian}.} \bibinfo{year}{2010}\natexlab{}.
\newblock \showarticletitle{Technique of Image Retrieval Based on Multi-label Image Annotation}. In \bibinfo{booktitle}{\emph{2010 Second International Conference on Multimedia and Information Technology}}, Vol.~\bibinfo{volume}{2}. \bibinfo{pages}{10--13}.
\newblock
\urldef\tempurl%
\url{https://doi.org/10.1109/MMIT.2010.34}
\showDOI{\tempurl}


\bibitem[Li et~al\mbox{.}(2015)]%
        {li2015gated}
\bibfield{author}{\bibinfo{person}{Yujia Li}, \bibinfo{person}{Daniel Tarlow}, \bibinfo{person}{Marc Brockschmidt}, {and} \bibinfo{person}{Richard Zemel}.} \bibinfo{year}{2015}\natexlab{}.
\newblock \showarticletitle{Gated graph sequence neural networks}.
\newblock \bibinfo{journal}{\emph{arXiv preprint arXiv:1511.05493}} (\bibinfo{year}{2015}).
\newblock


\bibitem[Lin et~al\mbox{.}(2014)]%
        {lin2014coco}
\bibfield{author}{\bibinfo{person}{Tsung-Yi Lin}, \bibinfo{person}{Michael Maire}, \bibinfo{person}{Serge Belongie}, \bibinfo{person}{James Hays}, \bibinfo{person}{Pietro Perona}, \bibinfo{person}{Deva Ramanan}, \bibinfo{person}{Piotr Doll{\'a}r}, {and} \bibinfo{person}{C.~Lawrence Zitnick}.} \bibinfo{year}{2014}\natexlab{}.
\newblock \showarticletitle{Microsoft COCO: Common Objects in Context}. In \bibinfo{booktitle}{\emph{Computer Vision -- ECCV 2014}}, \bibfield{editor}{\bibinfo{person}{David Fleet}, \bibinfo{person}{Tomas Pajdla}, \bibinfo{person}{Bernt Schiele}, {and} \bibinfo{person}{Tinne Tuytelaars}} (Eds.). \bibinfo{publisher}{Springer International Publishing}, \bibinfo{address}{Cham}, \bibinfo{pages}{740--755}.
\newblock
\showISBNx{978-3-319-10602-1}


\bibitem[Lu et~al\mbox{.}(2022)]%
        {lu2022prompt}
\bibfield{author}{\bibinfo{person}{Yuning Lu}, \bibinfo{person}{Jianzhuang Liu}, \bibinfo{person}{Yonggang Zhang}, \bibinfo{person}{Yajing Liu}, {and} \bibinfo{person}{Xinmei Tian}.} \bibinfo{year}{2022}\natexlab{}.
\newblock \showarticletitle{Prompt distribution learning}. In \bibinfo{booktitle}{\emph{Proceedings of the IEEE/CVF Conference on Computer Vision and Pattern Recognition}}. \bibinfo{pages}{5206--5215}.
\newblock


\bibitem[Paszke et~al\mbox{.}(2019)]%
        {pytorch2019}
\bibfield{author}{\bibinfo{person}{Adam Paszke}, \bibinfo{person}{Sam Gross}, \bibinfo{person}{Francisco Massa}, \bibinfo{person}{Adam Lerer}, \bibinfo{person}{James Bradbury}, \bibinfo{person}{Gregory Chanan}, \bibinfo{person}{Trevor Killeen}, \bibinfo{person}{Zeming Lin}, \bibinfo{person}{Natalia Gimelshein}, \bibinfo{person}{Luca Antiga}, \bibinfo{person}{Alban Desmaison}, \bibinfo{person}{Andreas Kopf}, \bibinfo{person}{Edward Yang}, \bibinfo{person}{Zachary DeVito}, \bibinfo{person}{Martin Raison}, \bibinfo{person}{Alykhan Tejani}, \bibinfo{person}{Sasank Chilamkurthy}, \bibinfo{person}{Benoit Steiner}, \bibinfo{person}{Lu Fang}, \bibinfo{person}{Junjie Bai}, {and} \bibinfo{person}{Soumith Chintala}.} \bibinfo{year}{2019}\natexlab{}.
\newblock \showarticletitle{PyTorch: An Imperative Style, High-Performance Deep Learning Library}. In \bibinfo{booktitle}{\emph{Advances in Neural Information Processing Systems}}, \bibfield{editor}{\bibinfo{person}{H.~Wallach}, \bibinfo{person}{H.~Larochelle}, \bibinfo{person}{A.~Beygelzimer}, \bibinfo{person}{F.~d\textquotesingle Alch\'{e}-Buc}, \bibinfo{person}{E.~Fox}, {and} \bibinfo{person}{R.~Garnett}} (Eds.), Vol.~\bibinfo{volume}{32}. \bibinfo{publisher}{Curran Associates, Inc.}
\newblock
\urldef\tempurl%
\url{https://proceedings.neurips.cc/paper_files/paper/2019/file/bdbca288fee7f92f2bfa9f7012727740-Paper.pdf}
\showURL{%
\tempurl}


\bibitem[Pennington et~al\mbox{.}(2014)]%
        {pennington2014glove}
\bibfield{author}{\bibinfo{person}{Jeffrey Pennington}, \bibinfo{person}{Richard Socher}, {and} \bibinfo{person}{Christopher~D Manning}.} \bibinfo{year}{2014}\natexlab{}.
\newblock \showarticletitle{Glove: Global vectors for word representation}. In \bibinfo{booktitle}{\emph{Proceedings of the 2014 conference on empirical methods in natural language processing (EMNLP)}}. \bibinfo{pages}{1532--1543}.
\newblock


\bibitem[Pu et~al\mbox{.}(2022)]%
        {pu2022semantic}
\bibfield{author}{\bibinfo{person}{Tao Pu}, \bibinfo{person}{Tianshui Chen}, \bibinfo{person}{Hefeng Wu}, {and} \bibinfo{person}{Liang Lin}.} \bibinfo{year}{2022}\natexlab{}.
\newblock \showarticletitle{Semantic-aware representation blending for multi-label image recognition with partial labels}. In \bibinfo{booktitle}{\emph{Proceedings of the AAAI Conference on Artificial Intelligence}}, Vol.~\bibinfo{volume}{36}. \bibinfo{pages}{2091--2098}.
\newblock


\bibitem[Pu et~al\mbox{.}(2024)]%
        {pu2024dual}
\bibfield{author}{\bibinfo{person}{Tao Pu}, \bibinfo{person}{Tianshui Chen}, \bibinfo{person}{Hefeng Wu}, \bibinfo{person}{Yukai Shi}, \bibinfo{person}{Zhijing Yang}, {and} \bibinfo{person}{Liang Lin}.} \bibinfo{year}{2024}\natexlab{}.
\newblock \showarticletitle{Dual-perspective semantic-aware representation blending for multi-label image recognition with partial labels}.
\newblock \bibinfo{journal}{\emph{Expert Systems with Applications}}  \bibinfo{volume}{249} (\bibinfo{year}{2024}), \bibinfo{pages}{123526}.
\newblock
\showISSN{0957-4174}
\urldef\tempurl%
\url{https://doi.org/10.1016/j.eswa.2024.123526}
\showDOI{\tempurl}


\bibitem[Pu et~al\mbox{.}(2021)]%
        {pu2021expression}
\bibfield{author}{\bibinfo{person}{Tao Pu}, \bibinfo{person}{Tianshui Chen}, \bibinfo{person}{Yuan Xie}, \bibinfo{person}{Hefeng Wu}, {and} \bibinfo{person}{Liang Lin}.} \bibinfo{year}{2021}\natexlab{}.
\newblock \showarticletitle{AU-Expression Knowledge Constrained Representation Learning for Facial Expression Recognition}. In \bibinfo{booktitle}{\emph{2021 IEEE International Conference on Robotics and Automation (ICRA)}}. \bibinfo{pages}{11154--11161}.
\newblock
\urldef\tempurl%
\url{https://doi.org/10.1109/ICRA48506.2021.9561252}
\showDOI{\tempurl}


\bibitem[Qin et~al\mbox{.}(2023)]%
        {qin2023freesegopenvoca}
\bibfield{author}{\bibinfo{person}{Jie Qin}, \bibinfo{person}{Jie Wu}, \bibinfo{person}{Pengxiang Yan}, \bibinfo{person}{Ming Li}, \bibinfo{person}{Ren Yuxi}, \bibinfo{person}{Xuefeng Xiao}, \bibinfo{person}{Yitong Wang}, \bibinfo{person}{Rui Wang}, \bibinfo{person}{Shilei Wen}, \bibinfo{person}{Xin Pan}, {et~al\mbox{.}}} \bibinfo{year}{2023}\natexlab{}.
\newblock \showarticletitle{Freeseg: Unified, universal and open-vocabulary image segmentation}. In \bibinfo{booktitle}{\emph{Proceedings of the IEEE/CVF Conference on Computer Vision and Pattern Recognition}}. \bibinfo{pages}{19446--19455}.
\newblock


\bibitem[Radford et~al\mbox{.}(2021)]%
        {radford2021CLIP}
\bibfield{author}{\bibinfo{person}{Alec Radford}, \bibinfo{person}{Jong~Wook Kim}, \bibinfo{person}{Chris Hallacy}, \bibinfo{person}{Aditya Ramesh}, \bibinfo{person}{Gabriel Goh}, \bibinfo{person}{Sandhini Agarwal}, \bibinfo{person}{Girish Sastry}, \bibinfo{person}{Amanda Askell}, \bibinfo{person}{Pamela Mishkin}, \bibinfo{person}{Jack Clark}, \bibinfo{person}{Gretchen Krueger}, {and} \bibinfo{person}{Ilya Sutskever}.} \bibinfo{year}{2021}\natexlab{}.
\newblock \showarticletitle{Learning Transferable Visual Models From Natural Language Supervision}. In \bibinfo{booktitle}{\emph{Proceedings of the 38th International Conference on Machine Learning}} \emph{(\bibinfo{series}{Proceedings of Machine Learning Research}, Vol.~\bibinfo{volume}{139})}, \bibfield{editor}{\bibinfo{person}{Marina Meila} {and} \bibinfo{person}{Tong Zhang}} (Eds.). \bibinfo{publisher}{PMLR}, \bibinfo{pages}{8748--8763}.
\newblock
\urldef\tempurl%
\url{https://proceedings.mlr.press/v139/radford21a.html}
\showURL{%
\tempurl}


\bibitem[Ridnik et~al\mbox{.}(2021)]%
        {ridnik2021asymmetric}
\bibfield{author}{\bibinfo{person}{Tal Ridnik}, \bibinfo{person}{Emanuel Ben-Baruch}, \bibinfo{person}{Nadav Zamir}, \bibinfo{person}{Asaf Noy}, \bibinfo{person}{Itamar Friedman}, \bibinfo{person}{Matan Protter}, {and} \bibinfo{person}{Lihi Zelnik-Manor}.} \bibinfo{year}{2021}\natexlab{}.
\newblock \showarticletitle{Asymmetric loss for multi-label classification}. In \bibinfo{booktitle}{\emph{Proceedings of the IEEE/CVF international conference on computer vision}}. \bibinfo{pages}{82--91}.
\newblock


\bibitem[Wang et~al\mbox{.}(2021)]%
        {wang2021pico}
\bibfield{author}{\bibinfo{person}{Haobo Wang}, \bibinfo{person}{Ruixuan Xiao}, \bibinfo{person}{Yixuan Li}, \bibinfo{person}{Lei Feng}, \bibinfo{person}{Gang Niu}, \bibinfo{person}{Gang Chen}, {and} \bibinfo{person}{Junbo Zhao}.} \bibinfo{year}{2021}\natexlab{}.
\newblock \showarticletitle{Pico: Contrastive label disambiguation for partial label learning}. In \bibinfo{booktitle}{\emph{International Conference on Learning Representations}}.
\newblock


\bibitem[Wang et~al\mbox{.}(2016)]%
        {wang2016cnn-rnn}
\bibfield{author}{\bibinfo{person}{Jiang Wang}, \bibinfo{person}{Yi Yang}, \bibinfo{person}{Junhua Mao}, \bibinfo{person}{Zhiheng Huang}, \bibinfo{person}{Chang Huang}, {and} \bibinfo{person}{Wei Xu}.} \bibinfo{year}{2016}\natexlab{}.
\newblock \showarticletitle{Cnn-rnn: A unified framework for multi-label image classification}. In \bibinfo{booktitle}{\emph{Proceedings of the IEEE conference on computer vision and pattern recognition}}. \bibinfo{pages}{2285--2294}.
\newblock


\bibitem[Wang et~al\mbox{.}(2023)]%
        {wang2023saliency}
\bibfield{author}{\bibinfo{person}{Shouwen Wang}, \bibinfo{person}{Qian Wan}, \bibinfo{person}{Xiang Xiang}, {and} \bibinfo{person}{Zhigang Zeng}.} \bibinfo{year}{2023}\natexlab{}.
\newblock \showarticletitle{Saliency Regularization for Self-Training with Partial Annotations}. In \bibinfo{booktitle}{\emph{Proceedings of the IEEE/CVF International Conference on Computer Vision}}. \bibinfo{pages}{1611--1620}.
\newblock


\bibitem[Wang et~al\mbox{.}(2017)]%
        {wang2017recurrent}
\bibfield{author}{\bibinfo{person}{Zhouxia Wang}, \bibinfo{person}{Tianshui Chen}, \bibinfo{person}{Guanbin Li}, \bibinfo{person}{Ruijia Xu}, {and} \bibinfo{person}{Liang Lin}.} \bibinfo{year}{2017}\natexlab{}.
\newblock \showarticletitle{Multi-label image recognition by recurrently discovering attentional regions}. In \bibinfo{booktitle}{\emph{Proceedings of the IEEE international conference on computer vision}}. \bibinfo{pages}{464--472}.
\newblock


\bibitem[Wu et~al\mbox{.}(2019)]%
        {wu2019instance}
\bibfield{author}{\bibinfo{person}{Hefeng Wu}, \bibinfo{person}{Yafei Hu}, \bibinfo{person}{Keze Wang}, \bibinfo{person}{Hanhui Li}, \bibinfo{person}{Lin Nie}, {and} \bibinfo{person}{Hui Cheng}.} \bibinfo{year}{2019}\natexlab{}.
\newblock \showarticletitle{Instance-aware representation learning and association for online multi-person tracking}.
\newblock \bibinfo{journal}{\emph{Pattern Recognition}}  \bibinfo{volume}{94} (\bibinfo{year}{2019}), \bibinfo{pages}{25--34}.
\newblock


\bibitem[Ye et~al\mbox{.}(2020)]%
        {ye2020SA}
\bibfield{author}{\bibinfo{person}{Jin Ye}, \bibinfo{person}{Junjun He}, \bibinfo{person}{Xiaojiang Peng}, \bibinfo{person}{Wenhao Wu}, {and} \bibinfo{person}{Yu Qiao}.} \bibinfo{year}{2020}\natexlab{}.
\newblock \showarticletitle{Attention-driven dynamic graph convolutional network for multi-label image recognition}. In \bibinfo{booktitle}{\emph{Computer Vision--ECCV 2020: 16th European Conference, Glasgow, UK, August 23--28, 2020, Proceedings, Part XXI 16}}. Springer, \bibinfo{pages}{649--665}.
\newblock


\bibitem[Yuan et~al\mbox{.}(2023)]%
        {Chen2023GraphAttention}
\bibfield{author}{\bibinfo{person}{Jin Yuan}, \bibinfo{person}{Shikai Chen}, \bibinfo{person}{Yao Zhang}, \bibinfo{person}{Zhongchao Shi}, \bibinfo{person}{Xin Geng}, \bibinfo{person}{Jianping Fan}, {and} \bibinfo{person}{Yong Rui}.} \bibinfo{year}{2023}\natexlab{}.
\newblock \showarticletitle{Graph Attention Transformer Network for Multi-label Image Classification}.
\newblock \bibinfo{journal}{\emph{ACM Trans. Multimedia Comput. Commun. Appl.}} \bibinfo{volume}{19}, \bibinfo{number}{4}, Article \bibinfo{articleno}{150} (\bibinfo{date}{feb} \bibinfo{year}{2023}), \bibinfo{numpages}{16}~pages.
\newblock
\showISSN{1551-6857}
\urldef\tempurl%
\url{https://doi.org/10.1145/3578518}
\showDOI{\tempurl}


\bibitem[Zhang and Peng(2021)]%
        {zhang2021instance}
\bibfield{author}{\bibinfo{person}{Zhiwei Zhang} {and} \bibinfo{person}{Hanyu Peng}.} \bibinfo{year}{2021}\natexlab{}.
\newblock \showarticletitle{Instance-weighted central similarity for multi-label image retrieval}.
\newblock \bibinfo{journal}{\emph{arXiv preprint arXiv:2108.05274}} (\bibinfo{year}{2021}).
\newblock


\bibitem[Zheng et~al\mbox{.}(2014)]%
        {zheng2014context}
\bibfield{author}{\bibinfo{person}{Yong Zheng}, \bibinfo{person}{Bamshad Mobasher}, {and} \bibinfo{person}{Robin Burke}.} \bibinfo{year}{2014}\natexlab{}.
\newblock \showarticletitle{Context Recommendation Using Multi-label Classification}. In \bibinfo{booktitle}{\emph{2014 IEEE/WIC/ACM International Joint Conferences on Web Intelligence (WI) and Intelligent Agent Technologies (IAT)}}, Vol.~\bibinfo{volume}{2}. \bibinfo{pages}{288--295}.
\newblock
\urldef\tempurl%
\url{https://doi.org/10.1109/WI-IAT.2014.110}
\showDOI{\tempurl}


\bibitem[Zhong et~al\mbox{.}(2022)]%
        {zhong2022regionclip}
\bibfield{author}{\bibinfo{person}{Yiwu Zhong}, \bibinfo{person}{Jianwei Yang}, \bibinfo{person}{Pengchuan Zhang}, \bibinfo{person}{Chunyuan Li}, \bibinfo{person}{Noel Codella}, \bibinfo{person}{Liunian~Harold Li}, \bibinfo{person}{Luowei Zhou}, \bibinfo{person}{Xiyang Dai}, \bibinfo{person}{Lu Yuan}, \bibinfo{person}{Yin Li}, {et~al\mbox{.}}} \bibinfo{year}{2022}\natexlab{}.
\newblock \showarticletitle{Regionclip: Region-based language-image pretraining}. In \bibinfo{booktitle}{\emph{Proceedings of the IEEE/CVF conference on computer vision and pattern recognition}}. \bibinfo{pages}{16793--16803}.
\newblock


\bibitem[Zhou et~al\mbox{.}(2022a)]%
        {zhou2022cocoop}
\bibfield{author}{\bibinfo{person}{Kaiyang Zhou}, \bibinfo{person}{Jingkang Yang}, \bibinfo{person}{Chen~Change Loy}, {and} \bibinfo{person}{Ziwei Liu}.} \bibinfo{year}{2022}\natexlab{a}.
\newblock \showarticletitle{Conditional prompt learning for vision-language models}. In \bibinfo{booktitle}{\emph{Proceedings of the IEEE/CVF conference on computer vision and pattern recognition}}. \bibinfo{pages}{16816--16825}.
\newblock


\bibitem[Zhou et~al\mbox{.}(2022b)]%
        {zhou2022coop}
\bibfield{author}{\bibinfo{person}{Kaiyang Zhou}, \bibinfo{person}{Jingkang Yang}, \bibinfo{person}{Chen~Change Loy}, {and} \bibinfo{person}{Ziwei Liu}.} \bibinfo{year}{2022}\natexlab{b}.
\newblock \showarticletitle{Learning to prompt for vision-language models}.
\newblock \bibinfo{journal}{\emph{International Journal of Computer Vision}} \bibinfo{volume}{130}, \bibinfo{number}{9} (\bibinfo{year}{2022}), \bibinfo{pages}{2337--2348}.
\newblock


\bibitem[Zhou et~al\mbox{.}(2023a)]%
        {Zhou2023AttentionAug}
\bibfield{author}{\bibinfo{person}{Wei Zhou}, \bibinfo{person}{Yanke Hou}, \bibinfo{person}{Dihu Chen}, \bibinfo{person}{Haifeng Hu}, {and} \bibinfo{person}{Tao Su}.} \bibinfo{year}{2023}\natexlab{a}.
\newblock \showarticletitle{Attention-Augmented Memory Network for Image Multi-Label Classification}.
\newblock \bibinfo{journal}{\emph{ACM Trans. Multimedia Comput. Commun. Appl.}} \bibinfo{volume}{19}, \bibinfo{number}{3}, Article \bibinfo{articleno}{116} (\bibinfo{date}{feb} \bibinfo{year}{2023}), \bibinfo{numpages}{24}~pages.
\newblock
\showISSN{1551-6857}
\urldef\tempurl%
\url{https://doi.org/10.1145/3570166}
\showDOI{\tempurl}


\bibitem[Zhou et~al\mbox{.}(2023b)]%
        {Zhou2023Algin}
\bibfield{author}{\bibinfo{person}{Wei Zhou}, \bibinfo{person}{Zhiwu Xia}, \bibinfo{person}{Peng Dou}, \bibinfo{person}{Tao Su}, {and} \bibinfo{person}{Haifeng Hu}.} \bibinfo{year}{2023}\natexlab{b}.
\newblock \showarticletitle{Aligning Image Semantics and Label Concepts for Image Multi-Label Classification}.
\newblock \bibinfo{journal}{\emph{ACM Trans. Multimedia Comput. Commun. Appl.}} \bibinfo{volume}{19}, \bibinfo{number}{2}, Article \bibinfo{articleno}{75} (\bibinfo{date}{feb} \bibinfo{year}{2023}), \bibinfo{numpages}{23}~pages.
\newblock
\showISSN{1551-6857}
\urldef\tempurl%
\url{https://doi.org/10.1145/3550278}
\showDOI{\tempurl}


\bibitem[Zhou et~al\mbox{.}(2023c)]%
        {Zhou2023DoubleAttention}
\bibfield{author}{\bibinfo{person}{Wei Zhou}, \bibinfo{person}{Zhiwu Xia}, \bibinfo{person}{Peng Dou}, \bibinfo{person}{Tao Su}, {and} \bibinfo{person}{Haifeng Hu}.} \bibinfo{year}{2023}\natexlab{c}.
\newblock \showarticletitle{Double Attention Based on Graph Attention Network for Image Multi-Label Classification}.
\newblock \bibinfo{journal}{\emph{ACM Trans. Multimedia Comput. Commun. Appl.}} \bibinfo{volume}{19}, \bibinfo{number}{1}, Article \bibinfo{articleno}{18} (\bibinfo{date}{jan} \bibinfo{year}{2023}), \bibinfo{numpages}{23}~pages.
\newblock
\showISSN{1551-6857}
\urldef\tempurl%
\url{https://doi.org/10.1145/3519030}
\showDOI{\tempurl}


\end{thebibliography}










\end{document}